\documentclass[sigconf, nonacm]{acmart}

\usepackage{balance}

\def\BibTeX{{\rm B\kern-.05em{\sc i\kern-.025em b}\kern-.08emT\kern-.1667em\lower.7ex\hbox{E}\kern-.125emX}}

\copyrightyear{2019}
\acmYear{2019}
\setcopyright{acmlicensed}
\acmConference[KDD '19] {The 25th ACM SIGKDD Conference on Knowledge Discovery and Data Mining}{August 4--8, 2019}{Anchorage, AK, USA}
\acmBooktitle{The 25th ACM SIGKDD Conference on Knowledge Discovery and Data Mining (KDD'19), August 4--8, 2019, Anchorage, AK, USA}
\acmPrice{15.00}
\acmDOI{10.1145/3292500.3330724}
\acmISBN{978-1-4503-6201-6/19/08}

\usepackage[utf8]{inputenc} 
\usepackage[T1]{fontenc}    
\usepackage{hyperref}       
\usepackage{url}            
\usepackage{booktabs}       
\usepackage{amsfonts}       
\usepackage{nicefrac}       
\usepackage{microtype}      
\usepackage{amsmath}
\usepackage{graphicx}
\usepackage{algorithmic}
\usepackage{algorithm}
\usepackage{xcolor}
\usepackage{multirow}
\usepackage{subcaption}
\usepackage{enumitem}
\numberwithin{algorithm}{section}
\usepackage[export]{adjustbox}

\usepackage{letters}
\usepackage{operators}

\definecolor{sred}{RGB}{255,63,88}
\definecolor{mred}{RGB}{191,47,66}
\definecolor{lred}{RGB}{122,30,42}

\title{A Deep Value-network Based Approach for Multi-Driver Order Dispatching}

\usepackage[belowskip=2pt,aboveskip=2pt]{caption}

\begin{document}
%
\author{Xiaocheng Tang}
\author{Zhiwei (Tony) Qin, Fan Zhang}
 \affiliation{%
  \institution{AI Labs, Didi Chuxing}
 }
 \email{xiaochengtang, qinzhiwei, feynmanzhangfan@didiglobal.com}
\author{Zhaodong Wang}
\authornote{Work done during an internship at Didi Chuxing.}
 \affiliation{%
  \institution{Washington State University}
 }
 \email{zhaodong.wang@wsu.edu}
\author{Zhe Xu}
 \affiliation{%
  \institution{Didi Chuxing}
 }
 \email{xuzhejesse@didiglobal.com}
\author{Yintai Ma}
\authornotemark[1]
 \affiliation{%
  \institution{Northwestern University}
 }
 \email{yintaima2020@u.northwestern.edu}

\author{Hongtu Zhu, Jieping Ye}
 \affiliation{
  \institution{AI Labs, Didi Chuxing}
 }
 \email{zhuhongtu,yejieping@didiglobal.com}

\renewcommand{\shortauthors}{Tang, et al.}

\begin{abstract}
Recent works on ride-sharing order dispatching have highlighted the importance of taking into account both the spatial and temporal dynamics in the dispatching process for improving the transportation system efficiency.
At the same time, deep reinforcement learning has advanced to the point where it achieves superhuman performance in a number of fields.
In this work, we propose a deep reinforcement learning based solution for order dispatching and we conduct large scale online A/B tests on DiDi's ride-dispatching platform to show that the proposed method achieves significant improvement on both total driver income and user experience related metrics.
In particular, we model the ride dispatching problem as a Semi Markov Decision Process to account for the temporal aspect of the dispatching actions.
To improve the stability of the value iteration with nonlinear function approximators like neural networks, we propose Cerebellar Value Networks (CVNet) with a novel distributed state representation layer.
We further derive a regularized policy evaluation scheme for CVNet that penalizes large Lipschitz constant of the value network for additional robustness against adversarial perturbation and noises.
Finally, we adapt various transfer learning methods to CVNet for increased learning adaptability and efficiency across multiple cities.
We conduct extensive offline simulations based on real dispatching data as well as online AB tests through the DiDi's platform. Results show that CVNet consistently outperforms other recently proposed dispatching methods.
We finally show that the performance can be further improved through the efficient use of transfer learning.
\end{abstract}

\maketitle

\section{Introduction}

In recent years, the advent of large scale online ride hailing services such as Uber and DiDi Chuxing have substantially transformed the transportation landscape,
offering huge opportunities for improving the current transportation efficiency and leading to a surge of interest in numerous research fields such as driving route planning, demand prediction, fleet management and order dispatching
(see, e.g., \cite{moreira2013predicting,xin2010aircraft,zhang2017taxi,zhang2016control}). One of the key enablers of this revolution lies in the ubiquitous use of Internet connected mobile devices which collect data and provide computation in real time. How to make use of the real-time rich information to
bridge the once significant gap between supply and demand and
improve traffic congestion, however, remains an active applied research topic.

In this work we consider the problem of driver-passenger dispatching \cite{liao2003real,zhang2016control,zhang2017taxi,xu2018large,wang2018deep}. In a ride-sharing platform the platform must make decisions for assigning available drivers to nearby unassigned passengers over a large spatial decision-making region (e.g., a city).
An optimal decision-making policy requires taking into account both the spatial extent and the temporal dynamics of the dispatching process since such decisions can have long-term effects on the distribution of available drivers across the city.
Previous work \cite{liao2003real,zhang2016control} ignores the global optimality in both the spatial and temporal dimensions, e.g., either assign the nearest driver to a passenger in a local greedy manner or match them on a first-come-first-serve basis.
In \cite{zhang2017taxi} the dispatching process is formulated as a combinatorial optimization problem. While only optimizing over the current time step, \cite{zhang2017taxi} demonstrates that by accounting for the spatial optimality alone a higher success rate of global order matches can be achieved.
\begin{figure}[]
\begin{center}
    \hspace*{-0.0in}\adjincludegraphics[width=0.5\textwidth, trim={0 0 0 0}, clip]{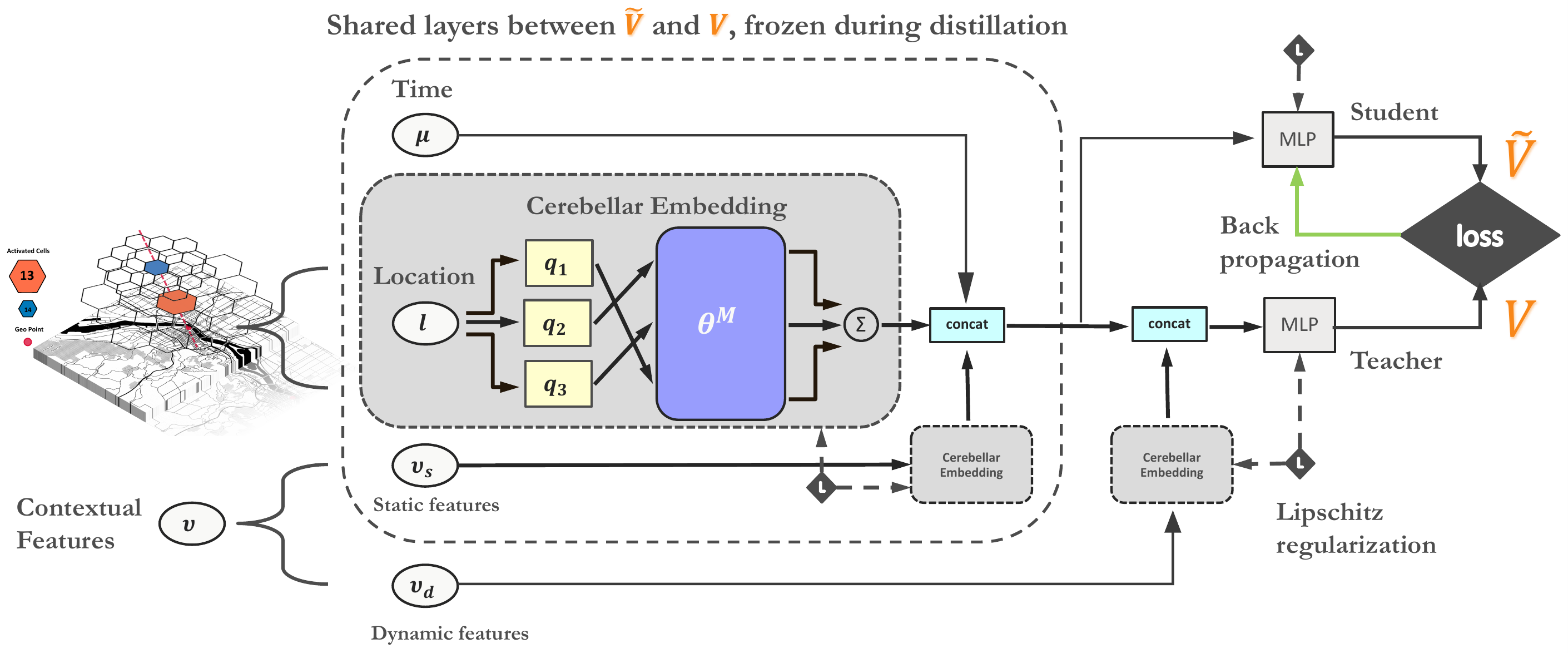}
    \caption{Feature marginalization and network structure of the Lipschitz-regularized dispatching value function $V$ and its distilled network $\tilde V$.
    }
    \label{fig:vnet-struct}
\end{center}
\vspace{-5pt}
\end{figure}

Two of the most recent work \cite{xu2018large,wang2018deep} stand out by explicitly considering the spatiotemporal optimality in the order dispatching model. The work in \cite{xu2018large} takes a learning and planning approach, where the offline learning step performs Temporal-Difference (TD) update in a discrete tabular spatiotemporal space using dynamic programming and the online planning step uses the tabular value from the learning step to compute real-time matching by solving a combinatorial optimization. While \cite{xu2018large} has successfully deployed the algorithm in the production system of Didi Chuxing and reported remarkable improvements over the baselines, the limitations of a tabular approach are obvious, e.g., unable to generalize beyond the historical training data and impossible to respond in real-time to different supply and demand conditions.
Another work in \cite{wang2018deep} adopts a deep reinforcement learning approach based on Q-learning, where a deep neural network is employed to estimate the state-action value function of the driver. The reinforcement learning agent is trained, from a single driver perspective, to maximize the total revenue throughout the day based on the historical data. To improve the sample complexity of reinforcement learning, a novel transfer learning method is also proposed for order dispatching to leverage knowledge transfer across multiple cities. \cite{wang2018deep} demonstrates the successful use of a deep neural network in a single driver's long term income maximization, but the approach is intrinsically inadequate to be used in a production dispatching system which demands coordinations among multiple agents.

Our contribution in this paper is a deep neural network based approach for order dispatching that achieves a significant improvement in large-scale online A/B test results through DiDi's ride-sharing platform.
The approach is proposed to maximize the long term total driver income for the multi-driver order dispatching environment.
In particular, we build upon and extend the learning and planning approach in \cite{xu2018large}.
Unlike \cite{wang2018deep,xu2018large} that formulate the dispatching problem as a standard MDP,
we base our algorithmic framework on a \emph{Semi-Markov Decision Process} (SMDP) formulation (e.g., see \cite{Sutton:1999fz,Bradtke1995}).
In fact, we show that the order dispatching decision process naturally involves choices among the temporally extended courses of action over a broad range of time scales, e.g., passenger assignments trigger driver transitions that take multiple time steps to finish depending on the duration of the trips.
The SMDP formulation accounts for such temporal effect of dispatching actions in a systematic way.
We show that this leads to an update rule that coincides with a heuristic for handling temporal extended actions used in \cite{xu2018large}.
In the learning phase the key of our method is the evaluation of a value function that estimates the \emph{expected} earning potential of a driver assuming that he or she follows the online policy (unknown) till the end of a day.
The use of a neural network to parameterize the value function allows us to extend beyond simply the spatiotemporal status and incorporates contextual information that reflects the real time supply/demand conditions. This provides more accurate and contextually sensitive driver income estimations. This value network is used as the input to the next planning phase, where a combinatorial problem similar to \cite{xu2018large} is solved to compute the actual dispatching decisions while resolving coordinations among multiple drivers and passengers.

The development of a contextual value network in the multi-driver order dispatching environment poses several challenges that desire novel solutions in both the learning and planning phases.
In fact, a straightforward substitution of function approximators for lookup tables in value iteration is not robust and may diverge, resulting in useless value estimates. This, of course, has been known for a long time and over the years numerous methods have been proposed to remedy the adverse effects.
A common theme is to build certain `invariance' either into the update process \cite{moore1995,van2016deep} or into the function approximator itself, e.g., constrain the changes in the output across a broad region of the input domain \cite{Yee92abstractionin,Sutton:1996vg}.

In this paper, we introduce Cerebellar Value Networks (CVNet), which is based on a type of memory-based neural networks
known as CMAC (Cerebellar Model Arithmetic Computer) \cite{cmac:albus}. CMAC uses multiple overlapping tilings/quantizations of the state space to produce feature representations. It is an associative memory with a built-in generalization and has been successfully applied to reinforcement learning problems
as a sparse-coarse-coded function approximator which is constrained to be robust
\cite{Sutton:1996vg,Yee92abstractionin}.
CVNet extends CMAC to a `wide' form, where tiles are associated with embedding vectors that are concatenated into the final feature representations. The representation capacity can thus be adjusted through the choices of embedding dimensions and the number of (overlapping) tiling/quantization functions. The use of CMAC ensures the built-in invariance against small perturbations as long as the current state is in the interior of all tiles. However, the value output can still suffer from abrupt changes if the perturbations result in a tile boundary crossing.
Following recent works on defending against adversarial attack \cite{szegedy2014,pmlr-v70-cisse17a} and improving the generalizability of neural networks \cite{Oberman2018}, we derive the Lipschitz of the \emph{cerebellar embedding} operator and formulate Lipschitz regularization for CVNet during training.
Together we show that CVNet with a regularized Lipschitz constant constitutes a robust and stable value function approximator with a strong generalization.

Finally, we overcome multiple practical issues and test the method both in simulator built with real-world data and in the large-scale production system of DiDi, which serves tens of millions of passengers and drivers on a daily basis. In particular, to account for the temporal variance often found in real-time contextual features, during training we employ a data argumentation procedure called \emph{context randomization};
to compute temporal difference in real time planning phase, we learn a separate CVNet with only the spatiotemporal inputs by knowledge distillation \cite{distill2015} from the original CVNet with the additional contextual inputs; and finally to facilitate learning across multiple cities, we adapt the transfer learning method proposed in \cite{wang2018deep} for single driver order dispatching and apply it to CVNet in a multi-driver environment.
We obtain state-of-arts improvement on key metrics including Total Driver Income (TDI) and user experience related metrics both in extensive simulations and in real world online AB test environment.

In what follows, we describe our SMDP formulation in Section~\ref{sec:smdp} and highlight the difference between a standard MDP. The Lipschitz regularized policy evaluation and CVNet structure are detailed in Section~\ref{sec:dispatch_policy_eval}, along with the context randomization technique we use for feature learning that generalizes. Section~\ref{sec:multi_driver_dispatching} discuss how to embed this neural network into a combinatorial problem for policy improvement in the online multi-agent environment with thousands of drivers.
In Section~\ref{sec:multi_city_transfer} we discuss the application of transfer learning in CVNet dispatching system. Experiment results are presented in Section~\ref{sec:experiments}. And finally Section~\ref{sec:conclusions} concludes the paper.

\section{A Semi-MDP Formulation}
\label{sec:smdp}
We model the system as a \emph{Markov decision process} endowed with a set of temporally extended actions. Such actions are also known as \emph{options} and the corresponding decision problem is known as a \emph{semi-Markov decision process}, or \emph{SMDP} (e.g., see \cite{Sutton:1999fz}).
In this framework a driver interacts episodically with an \emph{environment} at some discrete time scale, $t \in \Tcal := \{0,1,2,...,T\}$ until the \emph{terminal} time step $T$ is reached. On each time step, $t$, the driver perceives the state of the environment, described by the feature vector $s_t \in \Scal$, and on that basis chooses an option $o_t \in \Ocal_{s_t}$ that terminates in $s_{t'}$ where $t' = t+k_{o_t}$. As a response, the environment produces a numerical reward $r_{t+i}$ for each intermediate step, e.g., $i = 1,...,k_{o_t}$. We denote the expected rewards of the option model by $r_{st}^o := E\{r_{t+1} + \gamma r_{t+2} + ... + \gamma^{k_{o_t} - 1} r_{t+k_{o_t}} | s_t = s, o_t = o\}$ where $1 \geq \gamma > 0$ is the discount factor for the future reward. In the context of order dispatching, we highlight the following specifics:

\textbf{State}, $s_t$ consists of the geographical status of the driver $l_t$, the raw time stamp $\mu_t$ as well as the contextual feature vector given by $\upsilon_t$, i.e., $s_t := (l_t, \mu_t, \upsilon_t)$. The raw time stamp $\mu_t$ reflects the time scale in the real world and is independent of the discrete time $t$ that is defined for algorithmic purposes.
We use $\upsilon_t$ to represent the contextual feature vector at location $l_t$ and time $\mu_t$. We split contextual features into two categories, the dynamic features $\upsilon_{dt}$ such as real-time characteristics of supplies and demands within the vicinity of the given spatiotemporal point, and the static features $\upsilon_{st}$ containing static properties such as \emph{dayofweek}, driver service statics, holiday indicator, etc. When the discussion is focused on one particular time step we may ignore the subscript $t$ and directly write $\upsilon_d$ and $\upsilon_s$.

\textbf{Option}, denoted as $o_t$, represents the transition of the driver to a particular spatiotemporal status in the future, i.e., $o_t := l_{t + k_t}$ where $k_t = 0, 1, 2, ...$ is the duration of the transition which finishes once the driver reaches the destination. Executing option $o_t$ from state $s_t$ means starting the transition from origin $l_t$ to the destination specified by $o_t$.
This transition can happen due to either a trip assignment or an idle movement.
In the first case the option results in a nonzero reward, while in the latter case an idle option leads to a zero-reward transition that terminates at the place where the next trip option is activated. Note that different $o_t$ takes different time steps to finish and the time extension is often larger than 1, e.g., $k_t > 1$, which is one of the main differences from standard MDP.

\textbf{Reward}, $R_t$ is the total fee collected from a trip with a driver transition from $s_t$ to $s_{t'}$ by executing option $o_t$. $R_t$ is zero if the trip is generated from an idle movement. Conceptually $R_t$ can be considered as the sum of a sequence of immediate rewards received at each unit time step while executing the option $o_t$, e.g., $R_t = \sum_{i=1}^{k_t} r_{t+i}$. We use $\hat R_t$ to denote the discounted total reward over the duration of the option $o_t$ induced by the discount factor $\gamma$,
e.g., $\hat R_t = r_{t+1} + \gamma r_{t+2} + ... + \gamma^{k_t - 1} r_{t+k_t}$.

\textbf{Policy}, $\pi(o|s)$ specifies the probability of taking option $o$ in state $s$ regardless of the time step $t$. Executing $\pi$ in the environment generates a history of driver trajectories denoted as $\{\tau_i\}_{i \in \Hcal} := \{(s_{i0}, o_{i0}, r_{i1}, s_{i1}, o_{i1}, r_{i2}, ..., r_{iT_i}, s_{iT_i})\}_{i \in \Hcal}$ where $\Hcal$ denotes the index set of the historical driver trajectories. Associated with the policy $\pi$ is the \textbf{state value function} $V^{\pi}(s) := E\{\sum_{i=t+1}^T \gamma^{i-t-1} r_i | s_t = s\}$ which specifies the value of a state $s \in \Scal$ under the policy $\pi$ as the expected cumulative reward that the driver will gain starting from $s$ and following $\pi$ till the end of an episode.

\begin{figure}[]
\begin{center}
    \hspace*{-0.0in}\adjincludegraphics[width=0.5\textwidth, trim={0 0 0 0}, clip]{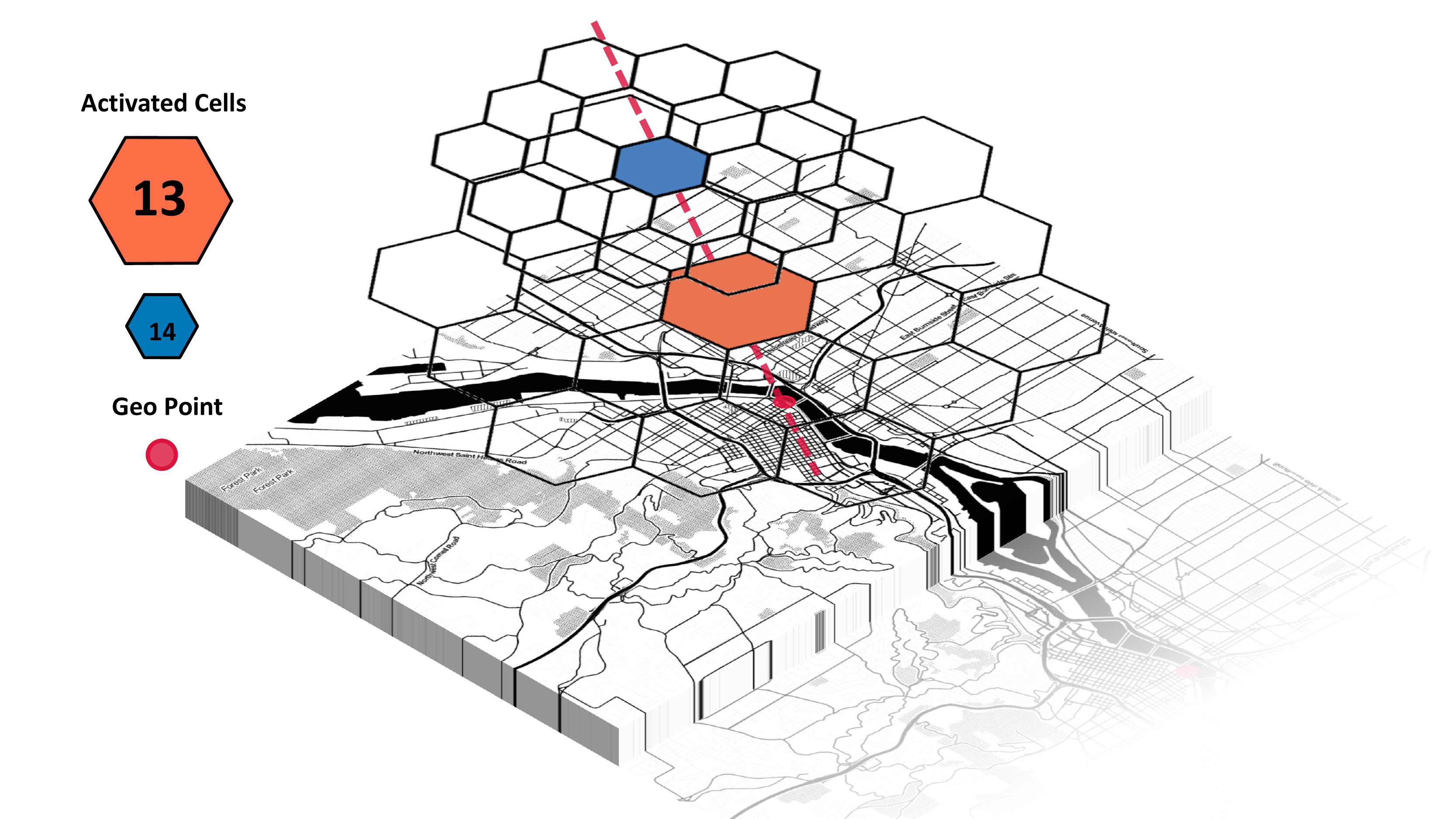}
    \caption{Coarse Coding with Hierarchical Hexagon Grid. The geo point (red) activates two grid cells (orange and blue). The final representation is the average of the two grid cells' embedding vectors.}
    \label{fig:cmac}
\end{center}
\end{figure}

\begin{figure}
\begin{center}
    \hspace*{-0.0in}\adjincludegraphics[width=0.4\textwidth, trim={0 0 0 0}, clip]{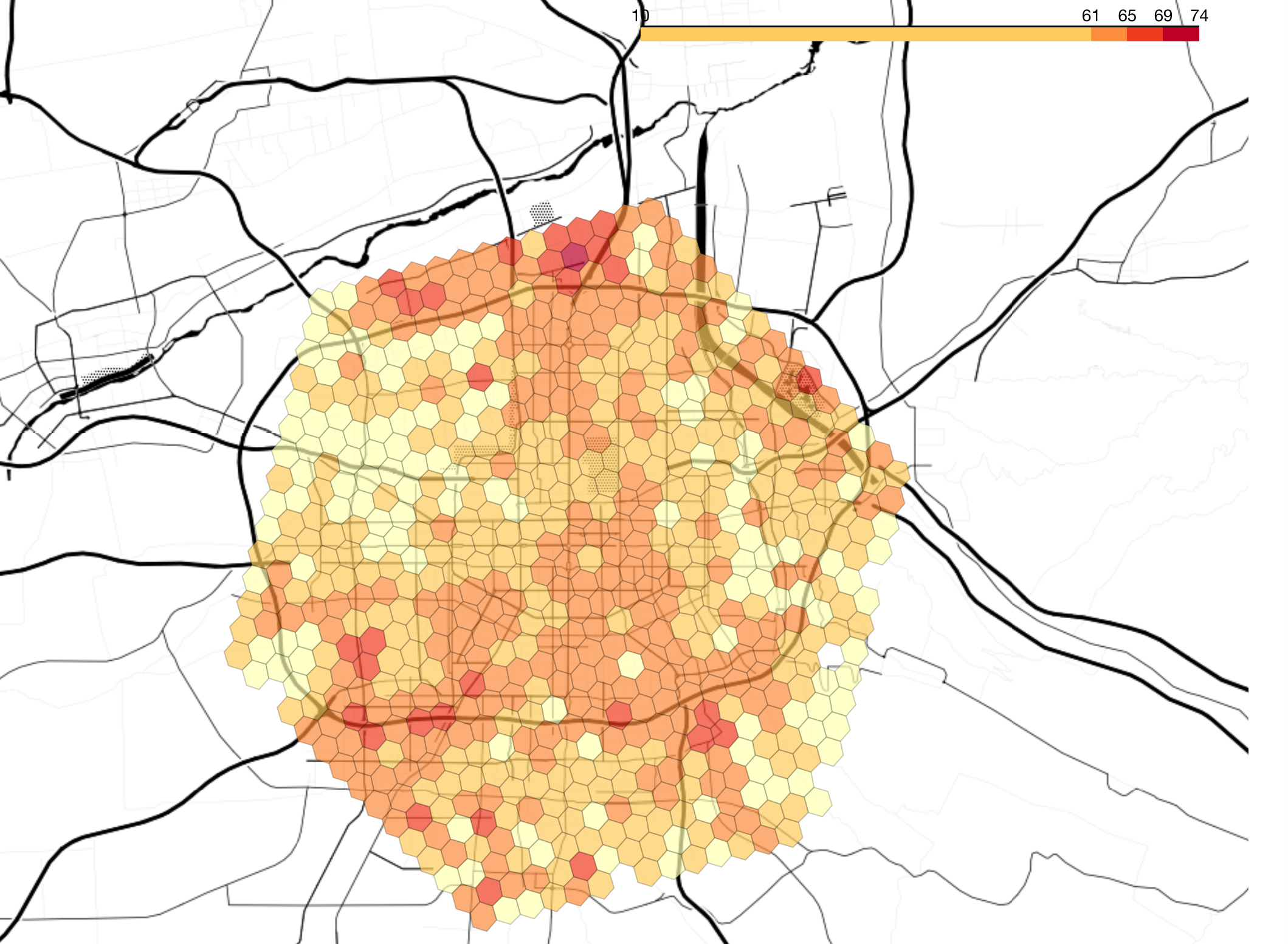}
    \caption{A visualization of CVNet output on a single layer of the hexagon grid system.}
    \label{fig:vnetvis}
\end{center}
\end{figure}
Given the above SMDP and the history trajectories $\Hcal$, our goal is to estimate the value of the underlying policy.
Similar to the standard MDP, we can write Bellman equations for general policies and options \cite{Sutton:1999fz},
\begin{align}
\label{equ:vpi_bellman}
    V^{\pi}(s)
    \notag &= E \{r_{t+1} + \cdots + \gamma^{k_{o_t} - 1} r_{t+k_{o_t}} + \gamma^{k_{o_t}} V^{\pi} (s_{t+k_{o_t}}) | s_t = s\} \\
    &= E \{r_{st}^o + \gamma^{k_{o_t}} V^{\pi} (s_{t+k_{o_t}}) | s_t = s\}
\end{align}
where $k_{o_t}$ is the duration of the option selected by $\pi$ at time $t$ and $r_{st}^o$ is the corresponding accumulative discounted reward received through the course of the option.

\paragraph{\textbf{Discussion}}
The Bellman equations \eqref{equ:vpi_bellman} can be used as update rules in dynamic-programming-like planning methods for finding the value function.
The main divergence from the standard MDP transition is that the update rules need to reflect the fact that the temporal extension from state to state spans different time horizons.
As an example, consider one transition from $s_t$ to $s_{t+k_t}$ resulted from executing option $o_t$. We can update the value function in this case as follows,
\begin{align*}
    V^{\kappa + 1}(s_t) \leftarrow r_{t+1} + \cdots + \gamma^{k_t - 1} r_{t+k_t} + \gamma^{k_t} V^{\kappa} (s_{t+k_t}).
\end{align*}
In the case of order dispatching, the total fee collected from the transition is $R_t$.
Assuming that $R_t$ is spread uniformly across the trip duration, we can then compute the discounted accumulative reward $\hat R_t$ as
\begin{align*}
    \hat R_t
    &= \frac{R_t}{k_t} + \gamma \frac{R_t}{k_t} +  \cdots + \gamma^{k_t - 1} \frac{R_t}{k_t} \\
    &= \frac{R_t(\gamma^{k_t} - 1)}{k_t(\gamma - 1)}, \quad \text{where } 0 < \gamma < 1, ~k_t \geq 1
\end{align*}
And the update rule for $V$ becomes
\begin{align}
\label{equ:v-target}
    V^{\kappa + 1}(s_t) \leftarrow \frac{R_t(\gamma^{k_t} - 1)}{k_t(\gamma - 1)} + \gamma^{k_t} V^{\kappa} (s_{t+k_t}).
\end{align}
Note that compared to a standard MDP update rule without reward discount $R_t + \gamma^{k_t} V^{\kappa} (s_{t+k_t})$, \eqref{equ:v-target} acts effectively like a smooth version of \emph{reward clipping} that is commonly used to improve performance in reinforcement learning \cite{mnih2015human}.

\section{Dispatching Policy Evaluation With Neural Networks}
\label{sec:dispatch_policy_eval}
We assume the online dispatching policy $\pi$ is unknown and the goal is to evaluate the value of the policy from the given historical trajectories data.
We use a neural network to approximate this value function based on the historical trajectories.
The network structure is illustrated in Figure~\ref{fig:vnet-struct}. Later we will discuss how to embed this neural network into a combinatorial problem for policy improvement in the online multi-agent environment with thousands of drivers.


\subsection{Cerebellar Embedding}
\label{subs:hierarchical_coarse_coded}
Learning a good state representation is usually the key step to solving a practical problem with neural networks. It is even more so for a large scale problem like order dispatching which requires the parse of complicated state information as the basis for long-term reasoning in a ever changing environment.
Here we propose a method called \emph{cerebellar embedding} that combines CMAC with embedding
to obtain a distributed state representation \cite{Hinton:1986:DR:104279.104287} that is generalizable, extensible and robust.
One way to view a CMAC is to consider a sparse, coarse-coded function approximator which uses multiple overlapping tilings of the state space to obtain a feature representation. Each input point to the CMAC activates as many tiles as the number of tilings used. The total number of tiles is referred to as the size of the \emph{conceptual memory}. The mapping from input points to tiles is done such that points close together in the input space have considerable overlap between their set of activated tiles. Each tile in the conceptual memory is associated with a weight in the \emph{actual memory} which is iteratively updated through training. And the output of CMAC is computed as the sum of the weights of the activated tiles. Note that the size of the \emph{actual memory} does not need to match that of the \emph{conceptual memory}. In fact, the so-called `hashing trick' \cite{Sutton:1996vg} is often employed to reduce the memory requirements -- a consistent random collapsing of a large set of tiles into a smaller one.

The cerebellar embedding extends CMACs by using an embedding matrix as the \emph{actual memory} and implements the mapping using a sparse representation.
In particular, the cerebellar embedding defines multiple quantization (or tiling) functions $\{q_1, ..., q_n\}$. Each function maps the continuous input to a unique string id indicating one discretized region of the state space such that $q_i(s) \neq q_j(s), \forall s, i \neq j$. The set of activated tiles for a given input $s$ is given by $\{q_i(s)\}_{i=1}^n$, and
the set of all such strings constitutes the \emph{conceptual memory}. The size of the \emph{actual memory} is denoted as $A$ which does not have to equal to the size of the \emph{conceptual memory}. Let $g(\cdot)$ denote a mapping function from the \emph{conceptual memory} to the range $0, 1, ..., A-1$. The \emph{perfect mapping} is when no conflict occurs, e.g., $g(q_i(s)) \neq g(q_j(s)), \forall i \neq j$.
Under the given set of quantization functions, we obtain the activation vector, denoted as $c(s) \in \Rmbb^A$, by iteratively adding 1 to the $g(q_i(s))$-th entry of $c(s)$ (initialized to $0$) for each $q_i$, e.g., $c_{g(q_i)}(s) \leftarrow c_{g(q_i)}(s) + 1, \forall i$. Hence $c(s)$ contains at most $n$ non-zero entries (exactly $n$ when it is \emph{perfect mappings}) and is a sparse vector since $n \ll A$.

Finally, we initiate a random embedding matrix $\theta^M \in \Rmbb^{A \times m}$ as the \emph{actual memory}. Each tile in the \emph{conceptual memory} is associated with a row in $\theta^M$ which is a dense $m$-dimensional vector.
The sparse activation vector $c(s)$ is multiplied by the embedding matrix, yielding the final dense representation of the input point $x$, i.e., $\frac{c(s)^T\theta^M}{n}$ where $n$ is the number of used quantization functions and the embedding matrix $\theta^M$ is iteratively updated during training. Note that the dot product $c(s)^T\theta^M$ grow linearly in magnitude with respect to the number of tilings so we scale it by $\frac{1}{n}$ to prevent diminishing gradients.

\subsubsection{Hierarchical Coarse-coding in the location space}
To quantize the geographical space, we use a hierarchical hexagon tiling system (illustrated in Figure \ref{fig:cmac}). Using a hexagon as the tile shape is beneficial since hexagons have only one distance between a hexagon centerpoint and its neighbors. The hexagon tiling system we use supports multiple resolutions, with each finer resolution having tiles with one seventh the area of the coarser resolution. Having such hierarchical quantization with different resolutions enables the information aggregation (and, in turn, the learning) to happen at different abstraction levels automatically adaptive to the nature of the geographical district, e.g., downtown, suburbs, community parks, etc.

\subsection{Robustness in Value Network}
\label{subs:continuity_in_vfunc}
Enforcing a robust state value dynamic with respect to the spatiotemporal status of the driver is critical in a production dispatching system. Dramatic changes or irregular value estimations will be further augmented due to either long chain of downstream tasks or simply large scale of the inputs, which can cause instability and abnormal behavior at the system level.
To obtain robustness against perturbations, mathematically we would like the output of the value function to be bounded, with respect to the $p$-norm of interest $||\cdot||_p$, by its input state for all state in $\Scal$, e.g.,
\begin{align}
\label{equ:lipschitz}
    ||V(s_1) - V(s_2)||_p \leq L_p ||s_1 - s_2||_p, \forall s_1, s_2 \in \Scal.
\end{align}
Here the value of $L_p$, known as the Lipschitz constant, represents the worst case variation of $V$ with respect to a change in its input $s$.
In this case we would like to regularize $L_p$ during training for a robust value function.

An upper bound for the Lipschitz constant of a neural network can be computed as the product of the Lipschitz constant of each individual layer of the network.
This is easy to show once we notice that neural networks can be expressed as a series of function compositions, e.g., $V(s) = (\nu_h \circ \nu_{h-1} \cdots \circ \nu_1)(s)$.
\begin{align}
\label{equ:lips-compose}
    L(V) \leq \Pi_{i=1}^h L(\nu_i)
\end{align}
Hence to control the neural network's global Lipschitz constant it is sufficient to regularize the Lipschitz for each individual layer.
The value network that we use, as depicted in Figure~\ref{fig:vnet-struct}, consists of both the cerebellar embedding layer and the multilayer perceptron. We now give the Lipschitz constants of these two layers as a function of their parameters.

\textbf{Multilayer Perceptron}:
Assume one linear layer followed by an ReLU activation. The Lipschitz of the ReLU operation is bounded by 1, e.g., $L_p^{relu} = 1$, since the maximum absolute subgradient of ReLU is 1.
For the linear layer, assuming it is parameterized by a weight matrix $\theta^l$ and a bias vector $b^l$, we can derive its Lipschitz constant as follows,
\begin{align*}
    &||\theta^l s_1 + b^l - (\theta^l s_2+b^l)||_p \leq L^l_p||s_1 - s_2||_p \\
    &\Rightarrow
    L^l_p \geq \frac{||\theta^l(s_1 - s_2)||_p}{||s_1-s_2||_p} \\
    &\Rightarrow
    L^l_p = \sup_{s \neq 0} \frac{||\theta^l s||_p}{||s||_p}, s = s_1 - s_2
\end{align*}
which is the operator norm of weight matrix $\theta^l$. When $p=1$ the Lipschitz constant of the linear layer $L^l_p$ is given by the maximum absolute column sum of the weight matrix; when $p=\infty$ it is the maximum absolute row sum and when $p=2$ it is the spectral norm of $\theta^l$ which can be approximated using the power method.

\textbf{Cerebellar Embedding}:
Recall that in Section~\ref{subs:hierarchical_coarse_coded} the embedding process can be expressed as a sparse dot product $\frac{c(s)^T\theta^M}{n}$ where $c(s)$ is a sparse vector with at most $n$ non-zero entries. Since this operation is linear in $c(s)$, the Lipschitz can be computed similarly as that of the linear layer. In this case it is the operator norm of the transpose of the embedding matrix $\theta^M$. Note that because quantizations are used, there will be a sudden change in the output value at the boundary of the quantization. This will not be an issue in practice as long as the scale of the change is controlled. That is if we regularize the operator norm of $\theta^M$.
In fact, note that the vector $c(s_1) - c(s_2)$ can have at most $2n$ non-zero entries for any $s_1, s_2$, e.g., when $s_1$ and $s_2$ have no overlap in the \emph{conceptual memory}. Hence the output of the cerebellar embedding layer is bounded as follows,
\begin{align*}
    &||c(x_1)^T\theta^M - c(x_2)^T\theta^M||_p / n \\
    &= ||(c(x_1) - c(x_2))^T\theta^M||_p / n
    \leq 2\max_i ||\theta^M_i||_p
\end{align*}
where $\theta^M_i$ is the $i$th row of $\theta^M$. When $p=1$, for example, $\max_i ||\theta^M_i||_1 = ||\theta^M||_{\infty}$ which is the infinity norm of the matrix $\theta^M$.


\subsection{Policy Evaluation}
\label{sub:policy_eval}

Given the semi-MDP defined in Section~\ref{sec:smdp}, we want to solve for the value function under the unknown dispatching policy $\pi$.
We collect the historical driver trajectories and divide it into a set of tuples with each representing one driver transition spending $k$ time steps from $s$ to $s'$ during which the driver receives a total trip fee $R$, i.e., $(s, R, s')$.
Training follows the Double-DQN structure \cite{van2016deep} for better training stability. The main value network is denoted as $V^{\pi}(s|\theta)$ where $\theta$ representing all trainable weights in the neural network,
and a target $V$-network $\hat V^{\pi}(s|\hat \theta)$, maintained and synchronized periodically with the main network $V^{\pi}(s|\theta)$, is used to evaluate the update rule as given in \eqref{equ:v-target}.
This update is converted into a loss to be minimized $\Lcal(\theta)$, most commonly the squared loss.
Following the discussions in Section \ref{subs:continuity_in_vfunc}, we
add a penalty term $\Rcal(\theta)$ on global Lipschitz constant to the loss and
introduce a penalty parameter $\lambda > 0$,
\begin{align}
\label{equ:reg-obj}
\begin{split}
    \min_{\theta}~
    &\Lcal(\theta) + \lambda \cdot \Rcal(\theta) := \\
    &\frac{1}{2}\{V^{\pi}(s|\theta) -
    (\frac{R(\gamma^{k} - 1)}{k(\gamma - 1)} + \gamma^{k} \hat V^{\pi}(s'|\hat \theta))\}^2
    + \lambda \cdot \sum_{i=1}^h L(\nu_i)
\end{split}
\end{align}
\textbf{Context Randomization:}
During training we augment each historical driver trajectory with contextual features $\{\upsilon_i\}$ extracted from the production logging system.
Contextual features, especially real-time supply/demand statistics, often come with high variance, e.g., it is common to notice a $\pm30$ minutes shift of the rush hour peak. Another issue is the scheduling bias in the logging system, e.g., logging triggered every 5 minutes, which can cause too many failed feature associations when matching using the exact value of the spatiotemporal states. To account for those bias in the training and to build temporal invariance into the system, we use \emph{context randomization} \eqref{equ:context-rand} in the augmentation process.
That is, instead of matching with the exact spatiotemporal status, we implement a procedure called \emph{hierarchical range query} $\Upsilon(\cdot)$,
which allows the specification of a range for the given query and
returns a set of contextual features within that range, i.e.,
$\Upsilon(l, \mu, rg) \subseteq \{\upsilon_i\}$
where $rg$ specify the query range for time $\mu$ such that all contextual features within $[\mu - rg, \mu + rg]$ are returned.

\begin{algorithm}
\caption{Regularized Policy Evaluation with Cerebellar Value Network (CVNet)}
\begin{algorithmic}[1]\label{alg:vnet}
\STATE Given:
historical driver trajectories $\{(s_{i,0}, o_{i,0}, r_{i,1}, s_{i,1}, o_{i,1}, r_{i,2}, ..., r_{i,T_i}, s_{i,T_i})\}_{i \in \Hcal}$ collected by executing a (unknown) policy $\pi$ in the environment.
\STATE Given: the \emph{hierarchical range query} function $\Upsilon(l, \mu, rg)$.
\STATE Given: $n$ cerebellar quantization functions $\{q_1, ..., q_n\}$, regularization parameter, max iterations, embedding memory size, embedding dimension, memory mapping function, discount factor, target update interval $\lambda, N, A, m, g(\cdot), \gamma, C > 0$.
\STATE Compute training data from the driver trajectories as a set of (state, reward, next state) tuples, e.g., $\{(s_{i,t}, R_{i,t}, s_{i,t+k_{i,t}})\}_{i \in \Hcal, t=0,...,T_i}$ where $k_{i, t}$ is the duration of the trip.
\STATE Initialize the state value network $V$ with random weights $\theta$ (including both the embedding weights $\theta^M \in \Rmbb^{A \times m}$ and the linear layer weights).
\STATE Initialize the target state value network $\hat{V}$ with weights $\hat{\theta}$.
\FOR{$\kappa = 1, 2, \cdots, N$}
  \STATE Sample a random mini-batch $\{(s_{i,t}, R_{i,t}, s_{i,t+k_{i,t}})\}$ from the training data.
  \STATE
  Sample $\upsilon$ randomly from the returned set of contextual features given query $l, \mu, rg$ and add it to the state $s$.
  \begin{align}
  \label{equ:context-rand}
  \begin{split}
    \upsilon_{i,t} &\in \Upsilon(l_{i,t}, \mu_{i,t}, rg), \\
    \upsilon_{i,t+k_{i,t}} &\in \Upsilon(l_{i,t+k_{i,t}}, \mu_{i,t+k_{i,t}}, rg)
  \end{split}
  \end{align}
  \STATE Transform the mini-batch into a (feature, label) format, e.g., $\{(x_i, y_i)\}$ where $x_i$ is $s_{i,t}$ and $y_i = \frac{R_{i,t}(\gamma^{k_{i,t}} - 1)}{k_{i,t}(\gamma - 1)} + \gamma^{k_{i,t}} \hat{V}(s_{i, t+k_{i,t}}) $
  \STATE Compute mini-batch gradient $\nabla_{|\{x_i, y_i\}} \Lcal(\theta) + \lambda \Rcal(\theta)$ according to \eqref{equ:reg-obj}
  \STATE Perform a gradient descent step on $\theta$ with $\nabla_{|\{x_i, y_i\}} \Lcal(\theta) + \lambda \Rcal(\theta)$.
  \IF{$\kappa \mod C = 0$}
    \STATE $\hat{\theta} \gets \theta$
  \ENDIF
\ENDFOR
\RETURN $V$
\end{algorithmic}
\end{algorithm}

\section{Planning With Multi-driver Dispatching} 
\label{sec:multi_driver_dispatching}

The production environment is intrinsically multi-agent with multiple drivers fulfilling passengers orders at the same time.
A matching problem \cite{xu2018large} is usually formulated at this stage to optimally assign the orders collected within a dispatching window to a set of drivers, while also avoiding assignment conflicts such as matching one order with multiple drivers.
A utility score $\rho_{ij}$ is used to indicate the value of matching each driver $i$ and order $j$ pair, and the objective of the matching problem is to maximize the total utilities of the assignments $\arg\max_{x \in \Ccal} \sum_{i=1}^{m}\sum_{j=1}^n \rho_{ij}x_{ij}$ where $\{x_{ij}\}$ are binary decision variables subject to a set of constraints $\Ccal$ to ensure the feasibility of the final assignment solution, e.g., each order is at most assigned to one driver, etc.
This problem can be solved by standard matching algorithms, such as the Hungarian Method (a.k.a. KM algorithm).


Similar to the work in \cite{xu2018large}, we use the \emph{Temporal Difference error} between order's destination state $s_j$ and driver's current state $s_i$ as the utility score $\rho_{ij}$.
Given the policy value function $V(s)$ as described above, this could be computed as below,
\begin{align}
\label{equ:rho-tderror}
    \rho_{ij} =
    R_{ij}\frac{(\gamma^{k_{ij}} - 1)}{k_{ij}(\gamma - 1)} + \gamma^{k_{ij}} V(s_j) - V(s_i) + \Omega \cdot U_{ij}
\end{align}
where $R_{ij}$ is the trip fee collected after the driver $i$ deliver order $j$; $k_{ij}$ is the time duration of the trip and $\gamma$ is the discount factor to account for the future uncertainty.
Aside from the long term driver income captured in the first part of \eqref{equ:rho-tderror}, we also add an additional term $\Omega \cdot U_{ij}, ~\Omega \geq 0$ where $U_{ij}$ characterizes the user experience from both the driver $i$ and the passenger $j$ so that we optimize not only the driver income but also the experience for both sides. As an example, setting $U_{ij}$ to be the negative of driver-passenger distance will have the effect of minimizing the waiting time for the passenger.

\subsection{Feature Marginalization via Distillation}
$V(s_j)$ in \eqref{equ:rho-tderror} represents the state value at order's destination.
The real time dynamic features $\upsilon_d$ at the order's destination, however, is not available until the driver actually finishes the trip.
In other words, we need a separate V-function that can evaluate the state value under the absence of those real time dynamic features. Let us call this V-function $\tilde V$. Given $V^{\pi}$, $\tilde V$ can be obtained through the marginalization over those features,
$\tilde V^{\pi} = E_{\upsilon_d}\{V^{\pi}(l, \mu, \upsilon_s, \upsilon_d)\}$.
Here, the contextual features $\upsilon$ are split into two groups, the static features $\upsilon_s$ and those dynamic features that require real-time computation $\upsilon_d$. The expectation is taken under the historical distribution of $\upsilon_d$, e.g., $p(\upsilon_d | l, \mu, \upsilon_s)$.

We make use of knowledge distillation to approximate this expectation, treating $V$ as the teacher network and training $\tilde V$ to mimic the output of $V$. Figure~\ref{fig:vnet-struct} illustrates the network structures of $\tilde V$ which is built on top of the structure of $V$.
$V$ and $\tilde V$ share the same state representation layers to encourage common knowledge transfer but distinguish from each other by having their own MLP and final output layers.
We use the full original training set for $V$ as the transfer set and evaluate $V$ on $l, \mu, \upsilon$ sampled from the transfer set to obtain the targets for $\tilde V$.
We activate distillation during the training of $V$ before each model checkpoint. The weights of $V$, including the shared weights, are frozen and only the MLP layers of $\tilde V$ are updated during distillation. We find that the distillation usually converges in less than 5 epochs and the distillation becomes much faster at later stage of the training as $V$ training also converges. So we anneal the number of distillation epochs in the beginning of the training. Afterwards we only run one epoch of updating $\tilde V$ for every model checkpoint. We find that this helps prevent overfitting while reducing the computation overhead to the training.


\section{Multi-city Transfer} 
\label{sec:multi_city_transfer}

Order dispatching can be naturally formulated as a multi-task learning problem with each task targeting at one particular regional area (e.g., a city). Training a single agent for all tasks may not scale well and can even raise many practical concerns in both deployment and model serving.
On the other hand, training each task independently is clearly suboptimal.
To efficiently scale CVNet to multiple cities, in this work we employ a method called CFPT (correlated-feature progressive transfer) proposed by \cite{wang2018deep} for the single driver dispatching environment. In CFPT the state space (and its corresponding structure) is split into two groups based on their adaptivity across different cities and transfered using a parallel progressive structure \cite{rusu2016progressive}. The idea is to maximize the knowledge transferred from those adaptive inputs like contextual features and time, while letting target training focus on the nonadaptive part such as the absolute GPS locations that are specifically tied to the task/city. We include the specific network structure and implementation details in the supplement material and report in Section~\ref{sec:experiments} experiment results comparing different transfer methods with CFPT. We find that the performance of CVNet can be further improved through the efficient use of knowledge transferred from another city.


\section{Experiments} 
\label{sec:experiments}


\begin{table}
  \caption{Stats of the training data consisting of one-month of driver trajectories and contextual features collected from three Chinese cities. Features are stored in a <key, value> format with key being the name, time and location.}
  \label{table:cities}
  \centering
  \begin{tabular}{||c c c c||}
    \hline
    City    & Region  & \#Transition  & \#Feature (in rows)     \\
    \hline
    A
    & Western
    & $2.72 \times 10^7$
    & $2.50 \times 10^7$ \\
    B
    & Southern
    & $2.98 \times 10^7$
    & $3.74 \times 10^7$ \\
    C
    & Northern
    & $1.90 \times 10^7$
    & $1.47 \times 10^7$ \\
    D
    & Eastern
    & $5.19 \times 10^6$
    & $3.52 \times 10^6$ \\
    \hline
  \end{tabular}
\end{table}

\begin{figure}
\centering
        \begin{subfigure}[b]{0.23\textwidth}
                \centering
                \includegraphics[width=\linewidth]{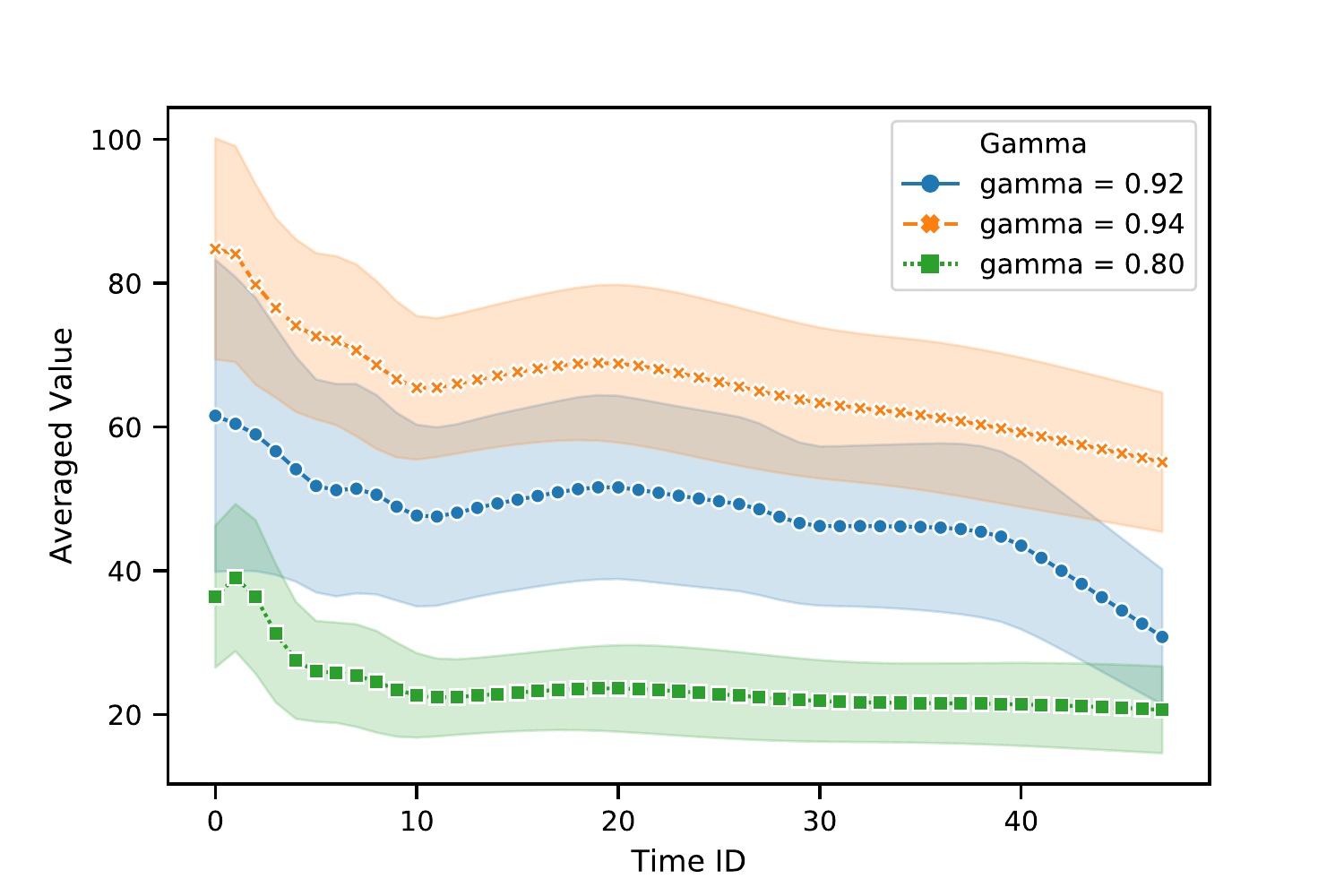}
                \caption{}
                \label{fig:gamma}
        \end{subfigure}%
        \begin{subfigure}[b]{0.24\textwidth}
                \centering
                \includegraphics[width=\linewidth]{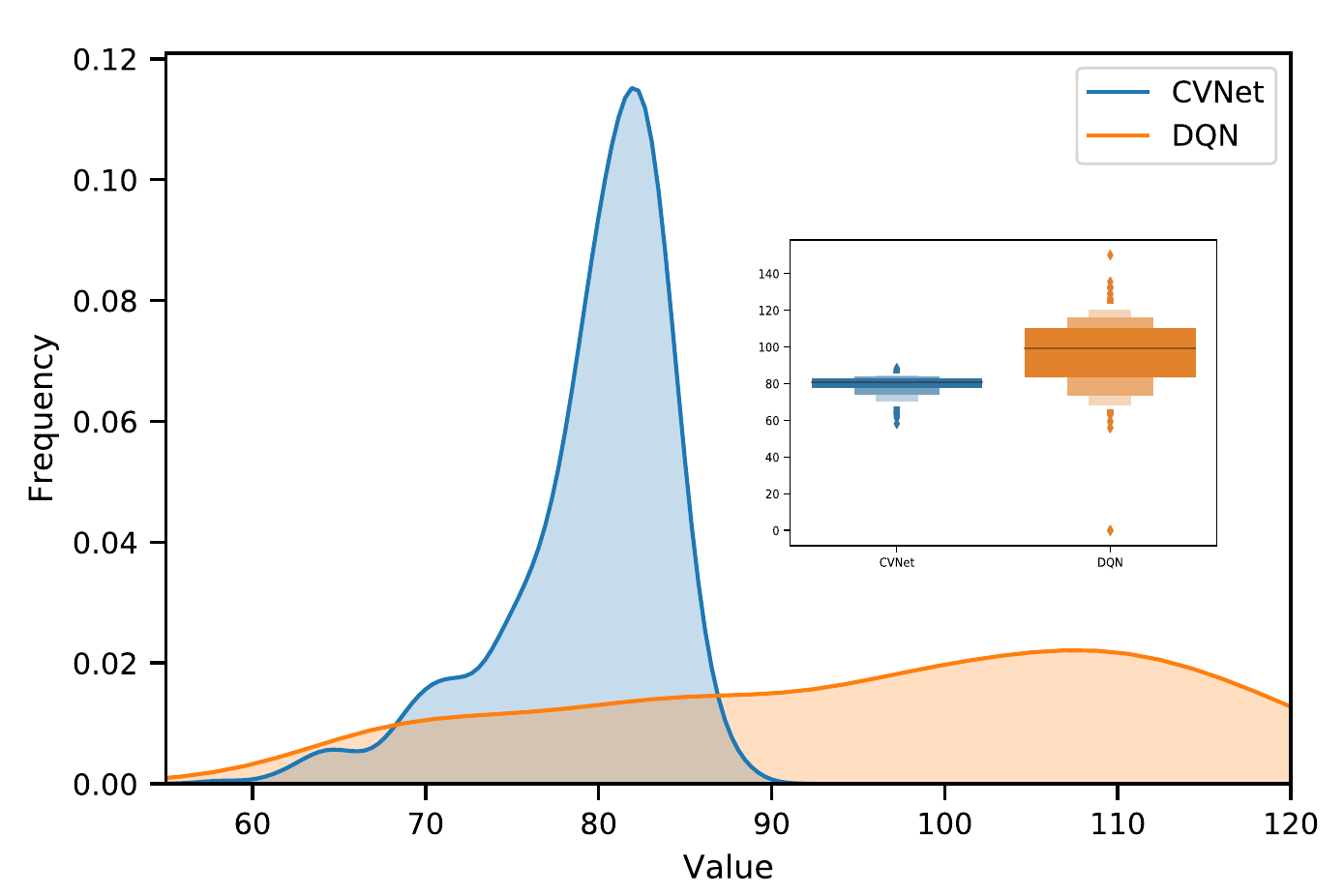}
                \caption{}
                \label{fig:histogram}
        \end{subfigure} \\%
        \begin{subfigure}[b]{0.23\textwidth}
                \centering
                \includegraphics[width=\linewidth]{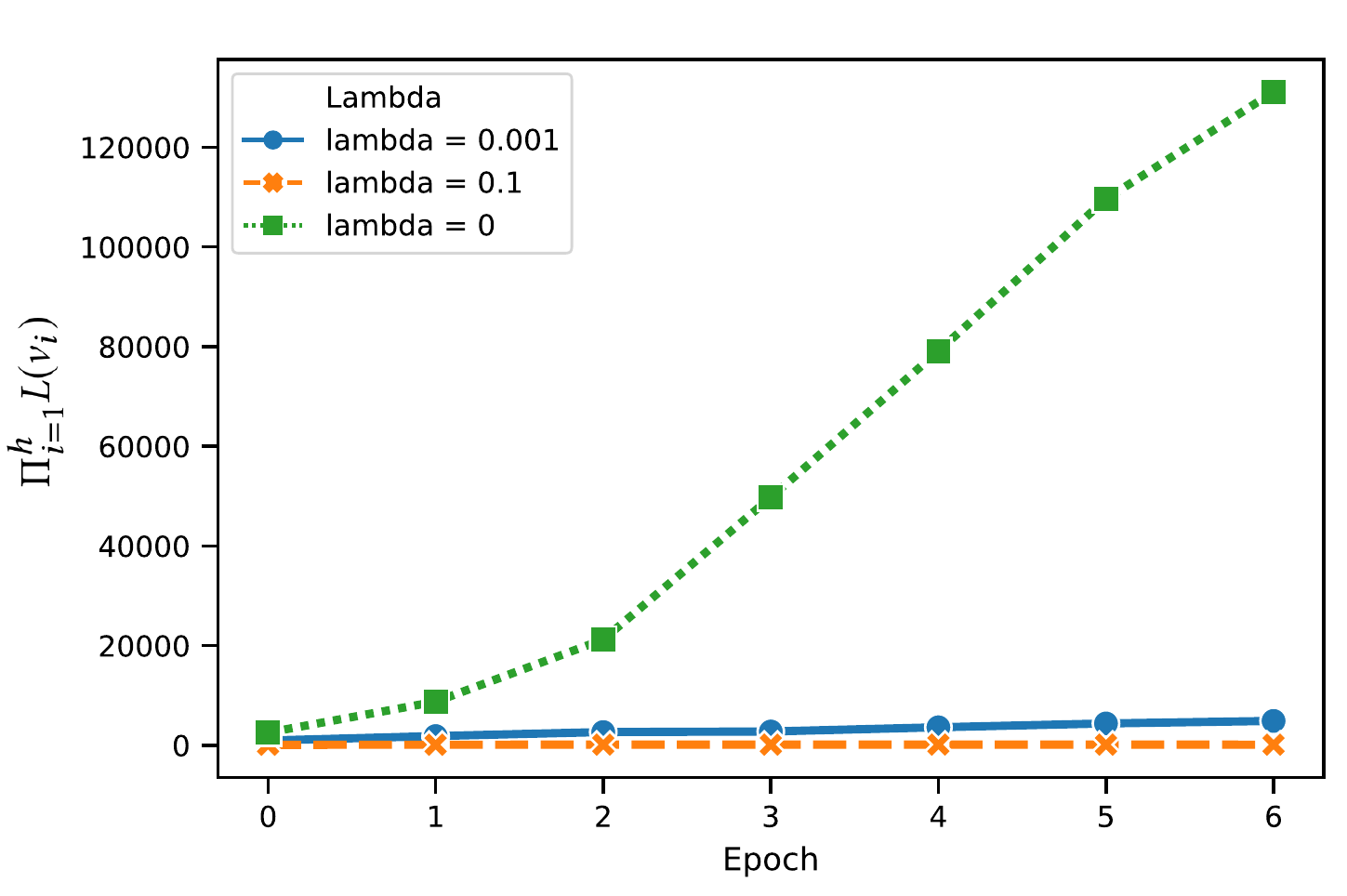}
                \caption{}
                \label{fig:lambda}
        \end{subfigure}
        \begin{subfigure}[b]{0.24\textwidth}
                \centering
                \includegraphics[width=\linewidth]{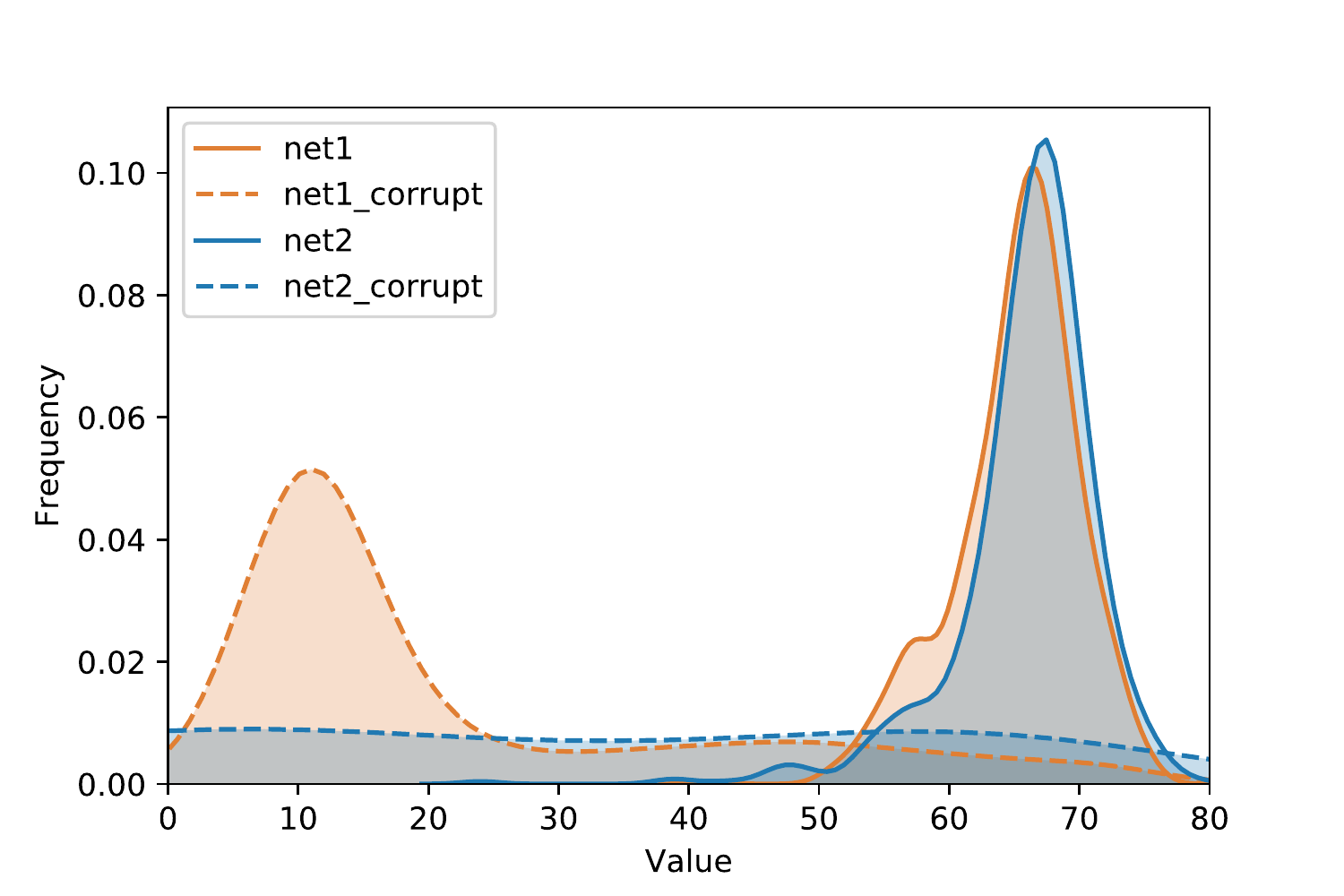}
                \caption{}
                \label{fig:corrupt}
        \end{subfigure}
        \caption{(a). Temporal patterns in the learned value network and how it reacts against the time discount factor $\gamma$; (b). Comparison of value distributions at a given time between DQN \cite{wang2018deep} and CVNet; (c). The change of global Lipschitz during training under different regularization $\lambda$; (d). Comparison of robustness of CVNet w/o Lipschitz Regularization (net1 is trained with $\lambda = 0.1$ and net2 is trained with $\lambda = 0$). }
    \label{fig:cvnet_parameters}
\end{figure}

\subsection{Characteristics of CVNet} 
\label{sub:character_cvnet}
We design various experiments to illustrate the robustness and spatiotemporal effect of CVNet. In this set of experiments we train CVNet on City A with embedding dimension set to 5 and without the use of contextual features.
The results are presented in Figure~\ref{fig:cvnet_parameters}.

Figure~\ref{fig:gamma} plots the change of the mean and standard deviation of $V$ value, evaluated on a fixed set of sampled locations from City A, against the time id of the day. Three curves are plotted, each with a different time discount factor $\gamma$ used during training.
We notice that for all curves the value of CVNet decreases towards zero as the time approaches the end of the day. This accords with
the definition (ref. Section~\ref{sec:smdp}) of CVNet.
Also note a higher averaged $V$ value as $\gamma$ increases to one (no discount).
In general, a small $\gamma$ induces a short-sighted strategy, e.g., the earnings over a one-hour period from now, while a large one encourages long-term behaviors. This also has an effect on the shape of the temporal patterns, as can be seen in the figure that for a small $\gamma = 0.8$ the value curve moves upwards temporarily during the morning rush hour period while the curves with large $\gamma$ approach zero in a more monotonic manner.

Figure~\ref{fig:lambda} demonstrates the effectiveness of Lipschitz regularization with the parameter $\lambda$ introduced in Section~\ref{sub:policy_eval}.
In Figure~\ref{fig:lambda} we plot, for different values of $\lambda$, the change of the bound on global Lipschitz \eqref{equ:lips-compose} as training progresses.
As expected the Lipschitz value explodes when there is no regularization $\lambda = 0$.
To see how the use of Lipschitz regularization improves the robustness and the training stability of CVNet, we employ a technique called \emph{weight corruption} which adds random noises to the weights of the hidden layers, analogous to, for example, the effect of a bad gradient descent step during training.
In this case, we corrupt the first layer of CVNet -- the embedding matrix $\theta^M$. We compute the output distribution against a fix sampled set of locations in City A and compare the change of this distribution before and after corruption.
As is shown in Figure~\ref{fig:corrupt}, the CVNet trained with Lipschitz regularization $\lambda = 0.1$ is much more robust against such noises compared to the one trained without Lipschitz regularization (both are corrupted with the same noise).
As we mentioned, any dramatic changes in the output distribution like the blue dashed curve shown in Figure~\ref{fig:corrupt} can have a detrimental effect on the training due to the recursive nature of the update rule, e.g., \eqref{equ:v-target}.

Finally, we compare the value distribution of CVNet with that of DQN \cite{wang2018deep} to show the "built-in" generalization of the \emph{cerebellar embedding}. The two methods are trained on the same dataset and evaluated at a given time step using the same set of popular latitude/longitude pairs from City A. It can be seen from the results in Figure~\ref{fig:histogram} that the value distribution of DQN not only exhibits a large variance, but also contains quite a few outliers that are clearly not correct (negative value).
Most of those abnormal values are from unseen locations that require generalization.
On the other hand, CVNet attains a distribution that is compact and robust against outliers.
With the use of cerebellar embedding, CVNet can generalize well to most unseen data points, since they are mapped to tiles whose embeddings are collaboratively learned during training.


\begin{figure}[]
\begin{center}
    \hspace*{-0.0in}\adjincludegraphics[width=0.48\textwidth, trim={0 0 0 0}, clip]{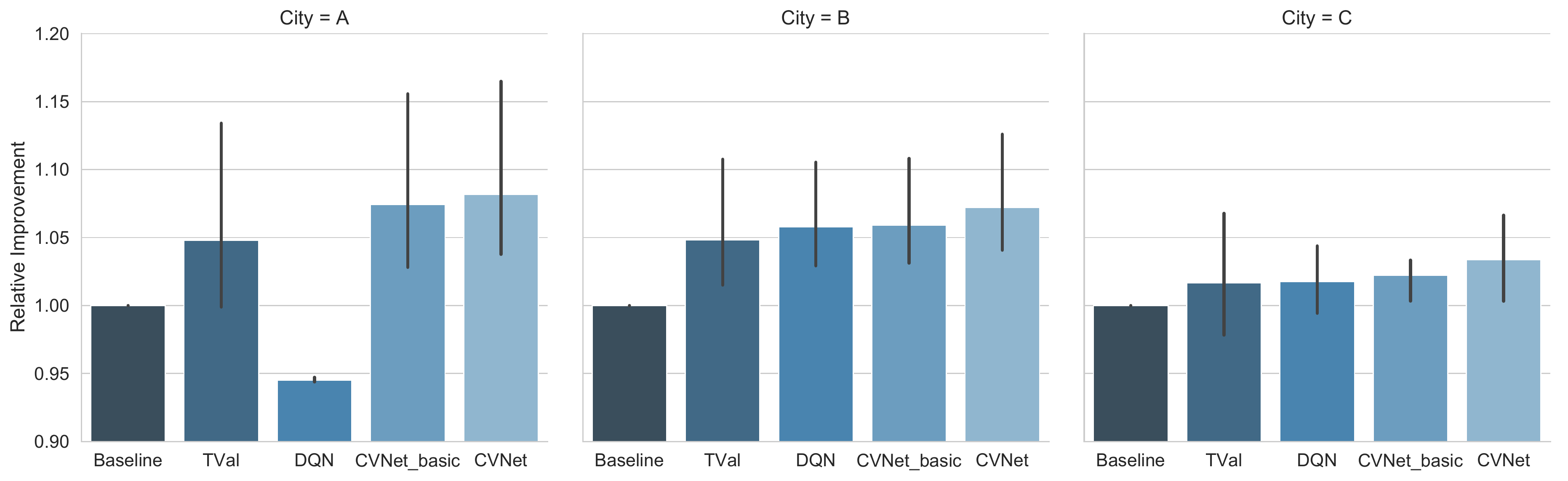}
    \caption{CVNet achieves the highest Total Driver Income (TDI) improvements compared to all other methods. }
    \label{fig:simbars}
\end{center}
\end{figure}

\begin{table}[]
\caption{Results from online AB test.}
\label{table:onlineab}
\begin{tabular}{|c|c|c|c|}
\hline
\multirow{2}{*}{} & \multicolumn{3}{c|}{Relative Improvement} \\ \cline{2-4}
                  &Answer rate (\%)
                  &Finish rate (\%)
                  &TDI (\%)\\ \hline
    City B    & \textbf{0.60}\ $\pm$\ 0.057 & \textbf{0.49}\ $\pm$\ 0.069  & \textbf{0.73}\ $\pm$\ 0.210  \\ \hline
    City C   & \textbf{1.16}\ $\pm$\ 0.062 & \textbf{1.11}\ $\pm$\ 0.083  & \textbf{0.93}\ $\pm$\ 0.198  \\ \hline
    City D  & \textbf{1.39}\ $\pm$\ 0.077  & \textbf{1.20}\ $\pm$\ 0.113  & \textbf{1.65}\ $\pm$\ 0.482  \\ \hline
\end{tabular}
\end{table}
\begin{figure*}[h!]
\centering
        \begin{subfigure}[b]{0.32\textwidth}
                \centering
                \includegraphics[width=\linewidth]{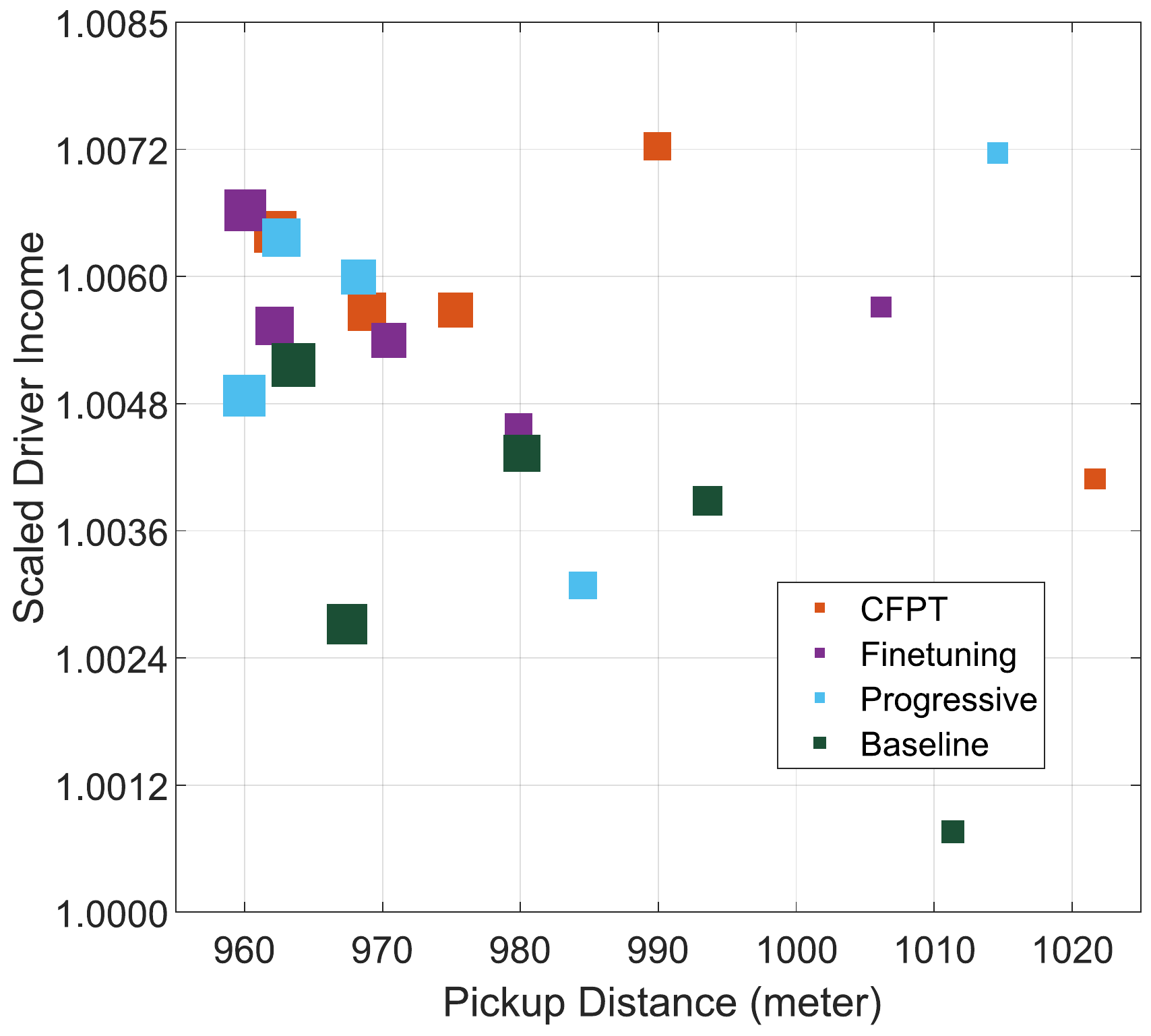}
                \caption{City B}
                \label{}
        \end{subfigure}%
        \vspace{6pt}
        \begin{subfigure}[b]{0.33\textwidth}
                \centering
                \includegraphics[width=\linewidth]{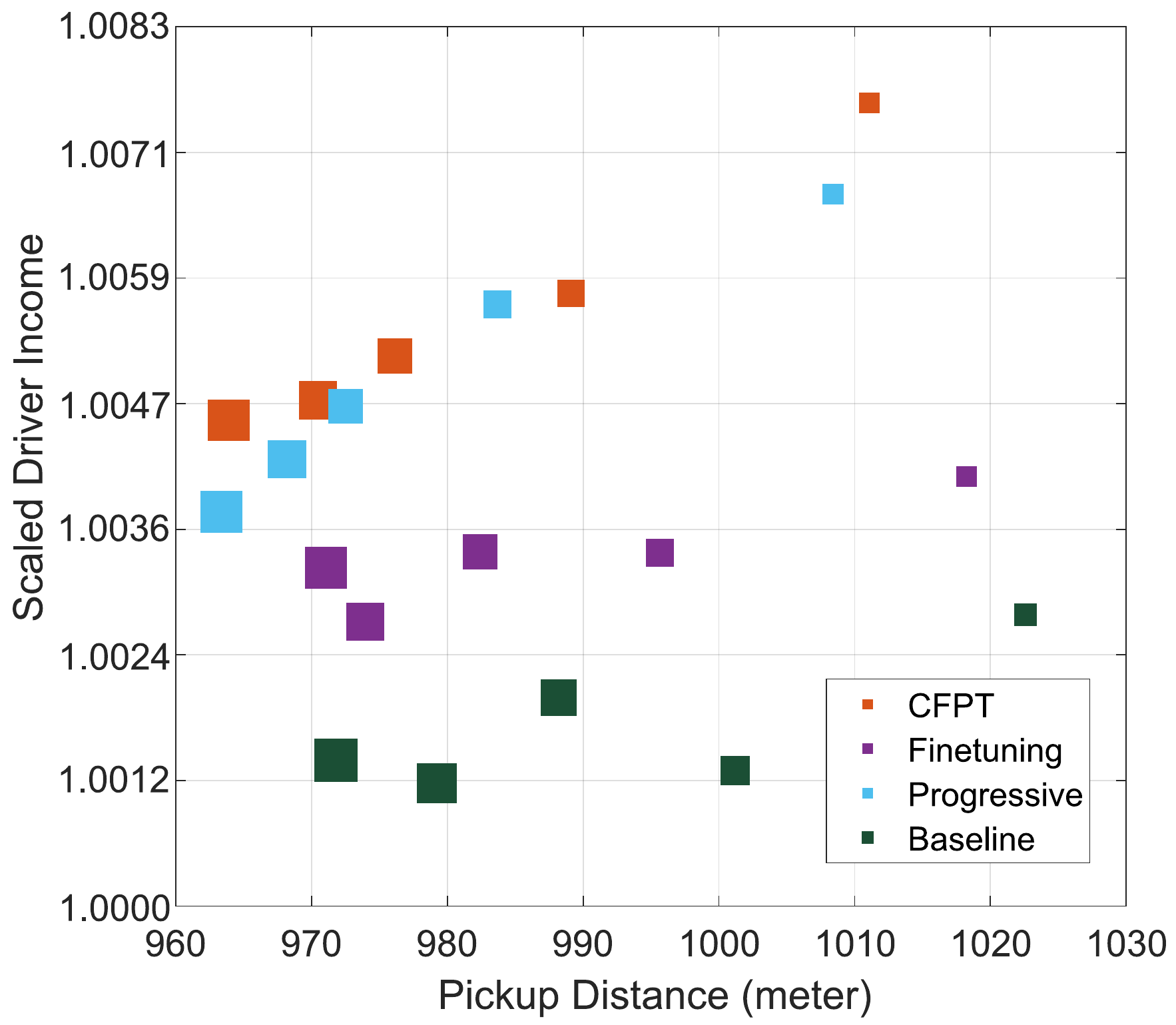}
                \caption{City C
                }
                \label{}
        \end{subfigure}%
        \begin{subfigure}[b]{0.33\textwidth}
                \centering
                \includegraphics[width=\linewidth]{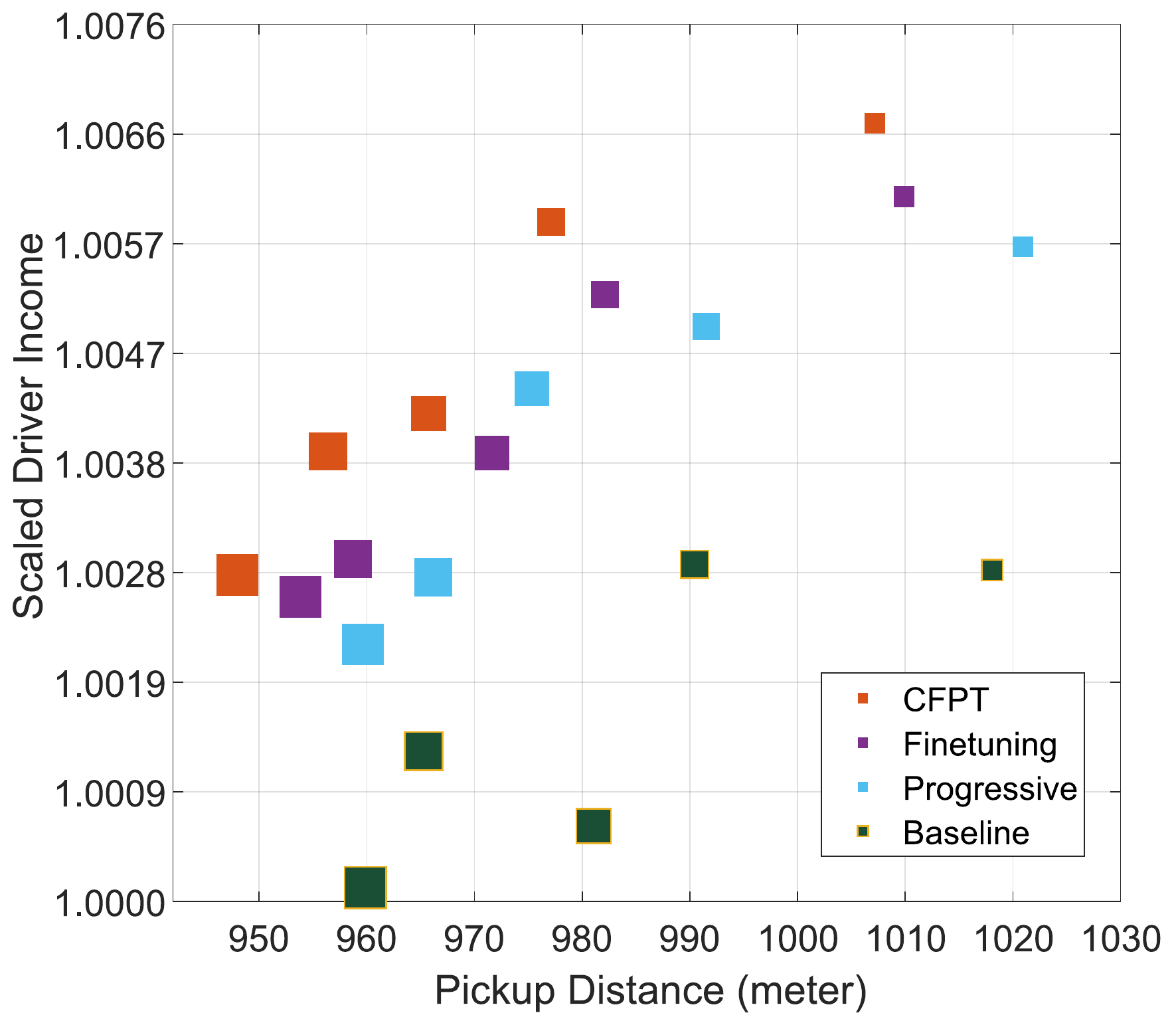}
                \caption{City D}
                \label{}
        \end{subfigure}
        \caption{Real data simulation result of the CVNet dispatching system, comparing the three transfer methods (from city A to B, C and D) with baseline. We have 5 different pick-up distance penalty levels here: a larger dot means a higher penalty, which will result in a smaller pickup distance (dots that are close to the left).  }
    \label{fig:transfer_gmv_pickup}
\end{figure*}

\subsection{Order Dispatching Experiments} 
\label{sub:order_dispatching_experiments}

Experiments in this section validate the capabilities of CVNet to improve the total driver income in a dynamic and complicated multi-driver environment.

\subsubsection{Results on Simulations With Real Data} 
\label{ssub:simulation_with_real_data}

We first use simulations with real data collected from DiDi's platform to validate CVNet, and more importantly as a level ground for comparing various order dispatching policies.

We give a brief descriptions below of the policies we compare with in the experiments.
Note that all policies compute the final dispatching decisions based on the assignment problem described in Section~\ref{sec:multi_driver_dispatching}. The variations come from the way the utility scores or edge weights on the bipartite graph $\rho_{ij}$ are computed, as we described below.

\begin{itemize}[leftmargin=*]
    \item \textbf{Baseline}. A myopic method that maximizes the current batch rate of dispatching, with $\rho_{ij}$ being the negative distance between driver-order pairs.
    \item \textbf{Tabular Value function (TVal)} \cite{xu2018large}. $\rho_{ij}$ is computed similarly as in \eqref{equ:rho-tderror} where the state evaluations only consider two variables, the time and location. The values are obtained by performing dynamic programming in a discrete tabular space. We ask authors of \cite{xu2018large} to kindly provide the production-ready set of values in a lookup table format for us to compare in this experiment.
    \item \textbf{DQN} \cite{wang2018deep}. A deep Q network is trained from a single driver's perspective. Here we compute $\rho_{ij}$ as the maximum Q value over all feasible action set, e.g., $\max_{a \in \tilde \Acal(s)} Q^*(s, a)$ where $s$ is approximated by the grid center and $\tilde \Acal(s)$ is computed from the set of historical trips originating from the vicinity of $s$. In this experiment we use the implementation and hyperparameters provided by the authors of \cite{wang2018deep} and train the deep Q networks using data listed in Table~\ref{table:cities}.
    \item \textbf{CVNet Basic}. The method proposed in this work with a simpler cerebellar embedding layer (flat hexagon grids with no hierarchy) and no contextual features. The final values are also stored in a lookup table format for use by the simulator. The main difference from TVal is then the use of a neural network as the function approximation. This serves the purpose of ablative analysis against CVNet.
    \item \textbf{CVNet}. The method proposed in this work with hexagon grid layers of three hierarchies. Four additional contextual features are used including one static feature, the dayofweek, and three dynamic ones including the trip query, order and empty driver count in the last one minute in the neighborhood of $s$.
\end{itemize}
Figure~\ref{fig:simbars} plots the simulations results averaged over three days including both weekdays and weekends. The Total Driver Income (TDI) of the Baseline is considered as one unit and the numbers from other methods are standardized accordingly. We notice in the results a consistent advantage of CVNet over the other methods across cities and days. In comparison with the Baseline, we notice that CVNet maintains an improvement ranging from $1\%$ to $15\%$ with an averaged improvement (across days) from $3\%$ to $8\%$.
CVNet Basic also performs quite well. While its advantage over TVal and DQN is not as significant as CVNet, it is considerably more robust and consistent in all situations.
Finally, note that DQN underperforms noticeably in City A. We suspect this is due to both the limitations of DQN's structure, e.g., no special treatment to ensure robustness, and the restriction of the single driver assumption, which make DQN vulnerable to both the noises in training and the changing dynamics in multi-driver environment.


\subsubsection{Results on the Real World} 
\label{ssub:real_world_results}
We conduct real-world experiments through DiDi's platform and report online A/B testing results in Table~\ref{table:onlineab}.
In the experiments \emph{CVNet Basic}\footnote{Due to online system restrictions at the time of the experiment.} is compared with the online production dispatching policy on three cities across China\footnote{City A is not available at the time of the experiment so we use City D instead.}. The experiments design and setup are similar to that used by \cite{xu2018large}.
Besides total driver income, we report two additional metrics including order answer rate and order finish rate.
Results are presented in Table~\ref{table:onlineab}. We notice from the results a consistent $0.5\% - 2\%$ improvement over the production baseline in all metrics across all three experiment cities.
An increase in order answer rate implies that over time the driver distribution has been optimized to align better with where orders might appear,
given that an order request will not be answered if there is no empty driver found in its neighborhood (usually a radius of 2km), and that the number of driver is not affected by the dispatching policy.
An increase in finish rate indicates that there are fewer trip cancellations after the orders are answered. Together they show that CVNet improves both the driver income and user experiences for the platform.



\subsection{Results on Transfer Across Cities} 
\label{sub:transfer_results}
In this set of experiments we study the effect of using transfer learning to scale CVNet across multiple cities. The experiments are conducted using the simulator. Four different transfer strategies are compared: CFPT \cite{wang2018deep}, Finetuning \cite{hinton2006reducing}, Progressive \cite{rusu2016progressive} and Baseline which simply trains a CVNet independently for each city without transfer.
City A is used as the source city and the rest three are used as the target city.
Figure~\ref{fig:transfer_gmv_pickup} shows the scaled Total Driver Income (TDI) and pickup distance under different pickup distance penalties -- $\Omega$ which is introduced in \eqref{equ:rho-tderror}. It can be observed from the figure that by altering the value of $\Omega$ we obtain a trade-off between TDI and pickup distance. By using transfer learning methods, especially CFPT, with CVNet, the trade-off curve is shifted significantly upwards. As a result, it is possible to attain a greater improvement on TDI while maintaining a short pickup distance.


\section{Conclusions and future work} 
\label{sec:conclusions}

This paper has proposed a deep reinforcement learning based solution for order dispatching. The method has been shown
to achieve significant improvement on both total driver income and user experience related metrics in large scale online A/B tests through DiDi's ride-dispatching platform.
First of all, a novel SMDP formulation has been proposed for the order dispatching problem to account for the temporally extended dispatching actions. Secondly, a new network structure, \emph{Cerebellar Value Networks} (CVNet), and a novel Lipschitz regularization scheme based on that structure have been proposed to ensure
both the robustness and the stability of the value iteration during policy evaluation.
Experiments using real data demonstrate that CVNet is robust against outliers and generalizes well to unseen data. Results on extensive simulations and online A/B testing have shown that CVNet outperforms all the other dispatching policies. Finally, we show that using transfer learning can further improve on the previous results and facilitate the scaling of CVNet across cities.

Our proposed approach consists of learning and planning two separate steps. Ideally we would like to combine them into one step that enables learning from end to end.
We would also like to explore ways of extending CVNet to other transportation applications like fleet management which has the similar goal of bridging the gap between supply and demand.
We leave these ideas as future directions of research.


\bibliographystyle{abbrv}
\balance
\bibliography{vnet}

\begin{thebibliography}{10}

\bibitem{cmac:albus}
J.~S. Albus.
\newblock {A theory of cerebellar function}.
\newblock {\em Mathematical Biosciences}, 10(1-2):25--61, 1971.

\bibitem{moore1995}
J.~A. Boyan and A.~W. Moore.
\newblock Generalization in reinforcement learning: Safely approximating the
  value function.
\newblock In G.~Tesauro, D.~S. Touretzky, and T.~K. Leen, editors, {\em
  Advances in Neural Information Processing Systems 7}, pages 369--376. MIT
  Press, 1995.

\bibitem{Bradtke1995}
S.~J. Bradtke and M.~O. Duff.
\newblock {Reinforcement learning methods for continuous-time Markov decision
  problems}.
\newblock {\em Advances in Neural Information Processing Systems (NIPS)}, 1995.

\bibitem{pmlr-v70-cisse17a}
M.~Cisse, P.~Bojanowski, E.~Grave, Y.~Dauphin, and N.~Usunier.
\newblock Parseval networks: Improving robustness to adversarial examples.
\newblock In {\em Proceedings of the 34th International Conference on Machine
  Learning}, volume~70, pages 854--863, International Convention Centre,
  Sydney, Australia, 06--11 Aug 2017.

\bibitem{distill2015}
G.~Hinton, O.~Vinyals, and J.~Dean.
\newblock Distilling the knowledge in a neural network.
\newblock In {\em NIPS Deep Learning and Representation Learning Workshop},
  2015.

\bibitem{Hinton:1986:DR:104279.104287}
G.~E. Hinton, J.~L. McClelland, and D.~E. Rumelhart.
\newblock Parallel distributed processing: Explorations in the microstructure
  of cognition, vol. 1.
\newblock chapter Distributed Representations, pages 77--109. MIT Press,
  Cambridge, MA, USA, 1986.

\bibitem{hinton2006reducing}
G.~E. Hinton and R.~R. Salakhutdinov.
\newblock Reducing the dimensionality of data with neural networks.
\newblock {\em science}, 313(5786):504--507, 2006.

\bibitem{liao2003real}
Z.~Liao.
\newblock Real-time taxi dispatching using global positioning systems.
\newblock {\em Communications of the ACM}, 46(5):81--83, 2003.

\bibitem{mnih2015human}
V.~Mnih, K.~Kavukcuoglu, D.~Silver, A.~A. Rusu, J.~Veness, M.~G. Bellemare,
  A.~Graves, M.~Riedmiller, A.~K. Fidjeland, G.~Ostrovski, et~al.
\newblock Human-level control through deep reinforcement learning.
\newblock {\em Nature}, 518(7540):529--533, 2015.

\bibitem{moreira2013predicting}
L.~Moreira-Matias, J.~Gama, M.-M.~J. Ferreira, Michel, and L.~Damas.
\newblock On predicting the taxi-passenger demand: A real-time approach.
\newblock In {\em Portuguese Conference on Artificial Intelligence}, pages
  54--65. Springer, 2013.

\bibitem{Oberman2018}
A.~M. Oberman and J.~Calder.
\newblock {Lipschitz regularized Deep Neural Networks converge and generalize}.
\newblock {\em arxiv preprint arXiv:1808.09540}, 2018.

\bibitem{rusu2016progressive}
A.~A. Rusu, N.~C. Rabinowitz, G.~Desjardins, H.~Soyer, J.~Kirkpatrick,
  K.~Kavukcuoglu, R.~Pascanu, and R.~Hadsell.
\newblock Progressive neural networks.
\newblock {\em arXiv preprint arXiv:1606.04671}, 2016.

\bibitem{Sutton:1996vg}
R.~S. Sutton.
\newblock {Generalization in reinforcement learning: Successful examples using
  sparse coarse coding}.
\newblock {\em Advances in Neural Information Processing Systems (NIPS)}, 1996.

\bibitem{Sutton:1999fz}
R.~S. Sutton, D.~Precup, and S.~Singh.
\newblock {Between MDPs and semi-MDPs: A framework for temporal abstraction in
  reinforcement learning}.
\newblock {\em Artificial Intelligence}, 112(1-2):181--211, Aug. 1999.

\bibitem{szegedy2014}
C.~Szegedy, W.~Zaremba, I.~Sutskever, J.~Bruna, D.~Erhan, I.~Goodfellow, and
  R.~Fergus.
\newblock Intriguing properties of neural networks.
\newblock In {\em International Conference on Learning Representations}, 2014.

\bibitem{van2016deep}
H.~Van~Hasselt, A.~Guez, and D.~Silver.
\newblock Deep reinforcement learning with double q-learning.
\newblock In {\em AAAI}, pages 2094--2100, 2016.

\bibitem{wang2018deep}
Z.~Wang, Z.~Qin, X.~Tang, J.~Ye, and H.~Zhu.
\newblock Deep reinforcement learning with knowledge transfer for online rides
  order dispatching.
\newblock In {\em IEEE International Conference on Data Mining}. IEEE, 2018.

\bibitem{xin2010aircraft}
T.~Xin-min, W.~Yu-ting, and H.~Song-chen.
\newblock Aircraft taxi route planning for a-smgcs based on discrete event
  dynamic system modeling.
\newblock In {\em Computer Modeling and Simulation, 2010. ICCMS'10. Second
  International Conference on}, volume~1, pages 224--228. IEEE, 2010.

\bibitem{xu2018large}
Z.~Xu, Z.~Li, Q.~Guan, D.~Zhang, Q.~Li, J.~Nan, C.~Liu, W.~Bian, and J.~Ye.
\newblock Large-scale order dispatch in on-demand ride-hailing platforms: A
  learning and planning approach.
\newblock In {\em Proceedings of the 24th ACM SIGKDD International Conference
  on Knowledge Discovery \& Data Mining}, pages 905--913. ACM, 2018.

\bibitem{Yee92abstractionin}
R.~Yee.
\newblock Abstraction in control learning.
\newblock Technical report, Technical Report COINS 92-16, Univ. of
  Massachusetts, 1992.

\bibitem{zhang2017taxi}
L.~Zhang, T.~Hu, Y.~Min, G.~Wu, J.~Zhang, P.~Feng, P.~Gong, and J.~Ye.
\newblock A taxi order dispatch model based on combinatorial optimization.
\newblock In {\em Proceedings of the 23rd ACM SIGKDD International Conference
  on Knowledge Discovery and Data Mining}, pages 2151--2159. ACM, 2017.

\bibitem{zhang2016control}
R.~Zhang and M.~Pavone.
\newblock Control of robotic mobility-on-demand systems: a queueing-theoretical
  perspective.
\newblock {\em The International Journal of Robotics Research},
  35(1-3):186--203, 2016.

\end{thebibliography}

\clearpage
\appendix

\section{Training Configuration}

To train the CVNet used in the experiments, we employ 3 cerebellar quantization functions and use a memory size $A$ of 20000. The embedding dimension $m$ is chosen to be 50.
Following the cerebellar embedding layer are fully connected layers having [32, 128, 32] hidden units with ReLU activations.
We maintain a target network which is updated every 100K steps. We use a batch size of 32 and run training for 20 epochs, with each epoch being one pass through the whole dataset. We apply Adam optimizer with a constant step size $3e^{-4}$. The Lipschitz regularization parameter $\lambda$ is chosen to be $1e^{-4}$ since we find that a small $\lambda$ is already quite effective at bounding the Lipschitz, as demonstrated in Figure~\ref{fig:gamma}. For context randomization we use a range $rg$ of 30 minutes.

\begin{figure}
\begin{center}
    \hspace*{-0in}\adjincludegraphics[width=0.9\columnwidth, trim={0 0 0 0}, clip]{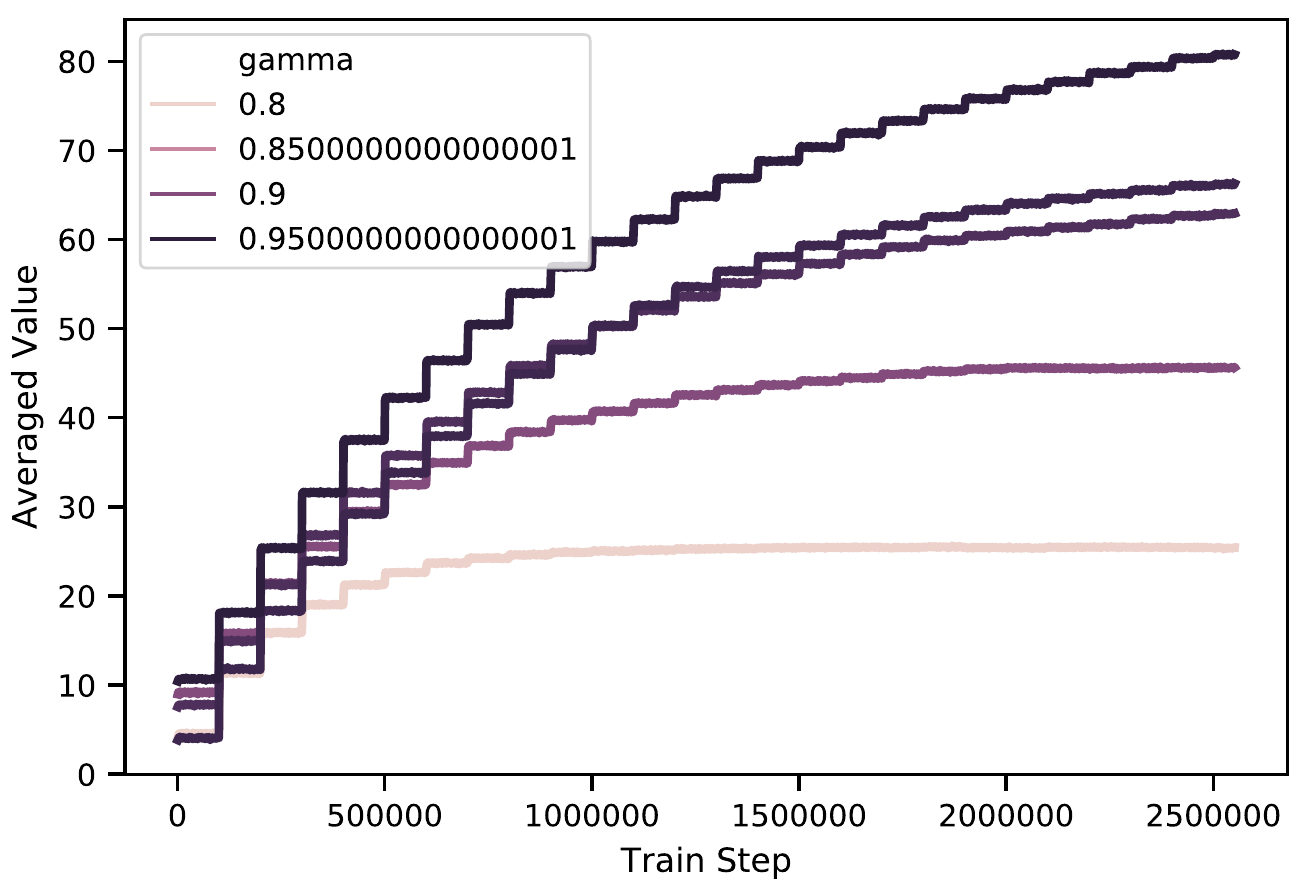}\hspace*{-0in}
    \caption{The average $V$ as training progresses. Each curve is generated by using a different discount factor $\gamma$ during policy evaluation, e.g., $V$ converges in roughly 1M steps when $\gamma = 0.8$.}
    \label{fig:gammatrain}
\end{center}
\end{figure}

Finally, we illustrate the training progress under different discount factors $\gamma$ in Figure~\ref{fig:gammatrain}.
During training we record the average $V$ for each input batch and we plot its change against the training steps. The average value the function $V$ converges to depends on the $\gamma$ being used. The value is smaller with a smaller $\gamma$, in which case the convergence also happens faster.
Note that a smaller $\gamma$ implies a more aggressive discounting of future values, hence a shorter lookahead horizon beyond which any rewards are close to zero once brought to the present. The training becomes easier in this case since it does not need to look far into the future.
In general $\gamma$ represents a trade-off between foresight and variance.
The $\gamma$ used in the experiments is chosen to be 0.92 which is determined by a randomized search based on out-of-sample simulations.

\section{Transfer Network Configuration}

In this supplement section, we will present the details of the three network models we implemented for knowledge transfer across multiple cities: fintuning~\cite{hinton2006reducing}, progressive network~\cite{rusu2016progressive}, and correlated feature progressive transfer (CFPT).

Details of these algorithms could be found in~\cite{wang2018deep}. Here we provide the network adaption due to the change of training algorithms (from Q network to state-value network).

\begin{figure}
\begin{center}
    \hspace*{-0in}\adjincludegraphics[width=0.8\columnwidth, trim={0 0 0 0}, clip]{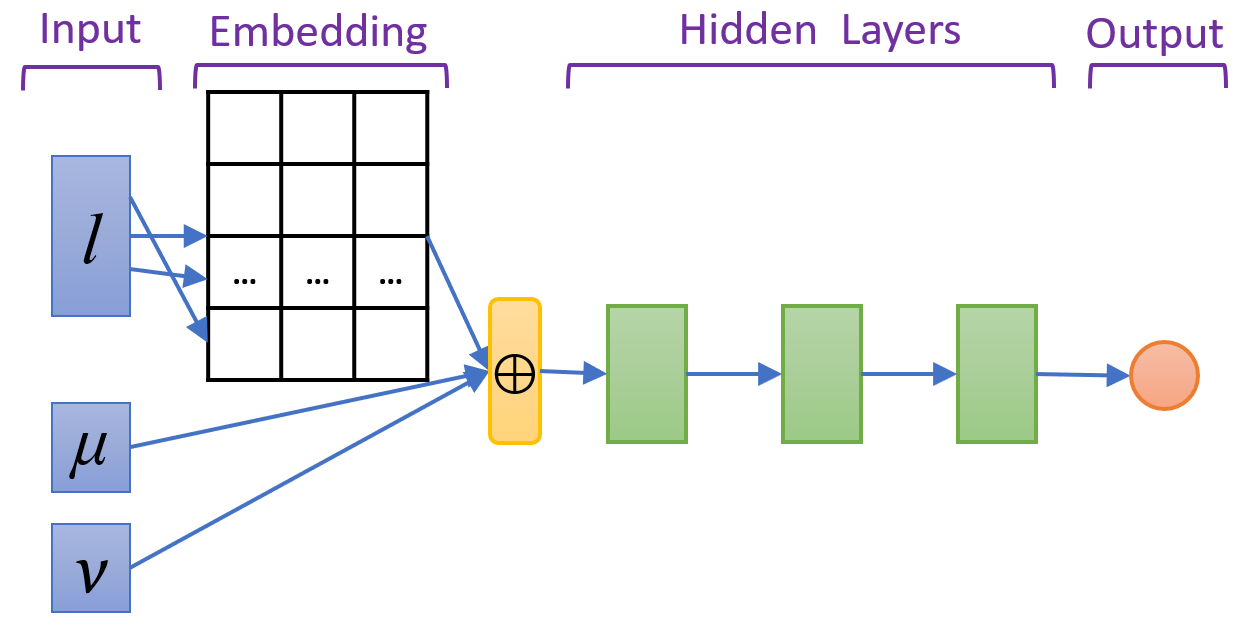}\hspace*{-0in}
    \caption{Network structure of CVNet}
    \label{fig:vnet_structure}
\end{center}
\end{figure}

\begin{figure}
\begin{center}
    \hspace*{-0in}\adjincludegraphics[width=0.9\columnwidth, trim={0 0 0 0}, clip]{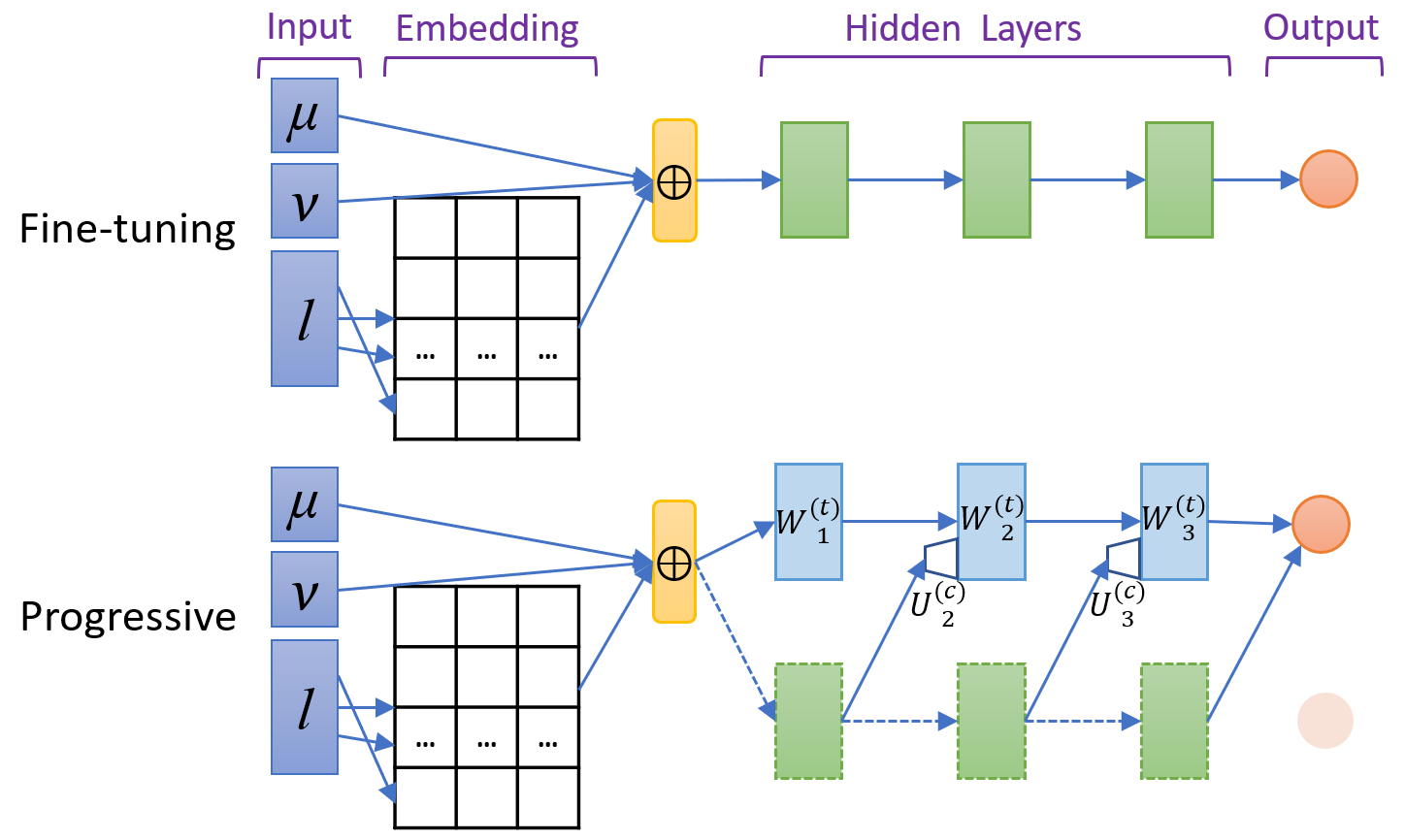}\hspace*{-0in}
    \caption{Structures of finetuning and progressive network. Green blocks are weights initialized with trained network from the source city. Frozen layers would keep the transferred weights during the target training.}
    \label{fig:fintuning_progressive_network}
\end{center}
\end{figure}

\textbf{Finetuning}: After training the network of the source city (using the original model in Figure~\ref{fig:vnet_structure}), we initialize the ``green blocks'' of the target city network with the trained weights as in Figure~\ref{fig:fintuning_progressive_network} and continue to train on the new dataset.

\textbf{Progressive Network}: Instead of directly initializing the target network, lateral connection is used to leverage the trained weights as in~\ref{fig:fintuning_progressive_network}. The parallel network (green blocks) remains the trained weights from the source city. The connection is defined as:
\begin{equation}
h_i^{(k)}=f \bigg( W_i^{(k)} h_{i-1}^{(k)}+ \sum_{j<k}U_i^{(j:k)}h_{i-1}^{(j)} \bigg),
\label{func:progressive_connection}
\end{equation}
where $h_{i}^{(t)}$ and $h_{i}^{(s)}$ denote the outputs of layer $i$ in the target network and the source network, correspondingly.
$W_i^{(t)}$ is the weight matrix of layer $i$ of the target network, and $U_i^{(c)}$ is the lateral connection weight matrix from the ``green blocks'' of the source tasks. $f(\cdot)$ is the activation function.

\begin{figure}
\begin{center}
    \hspace*{-0in}\adjincludegraphics[width=0.9\columnwidth, trim={0 0 0 0}, clip]{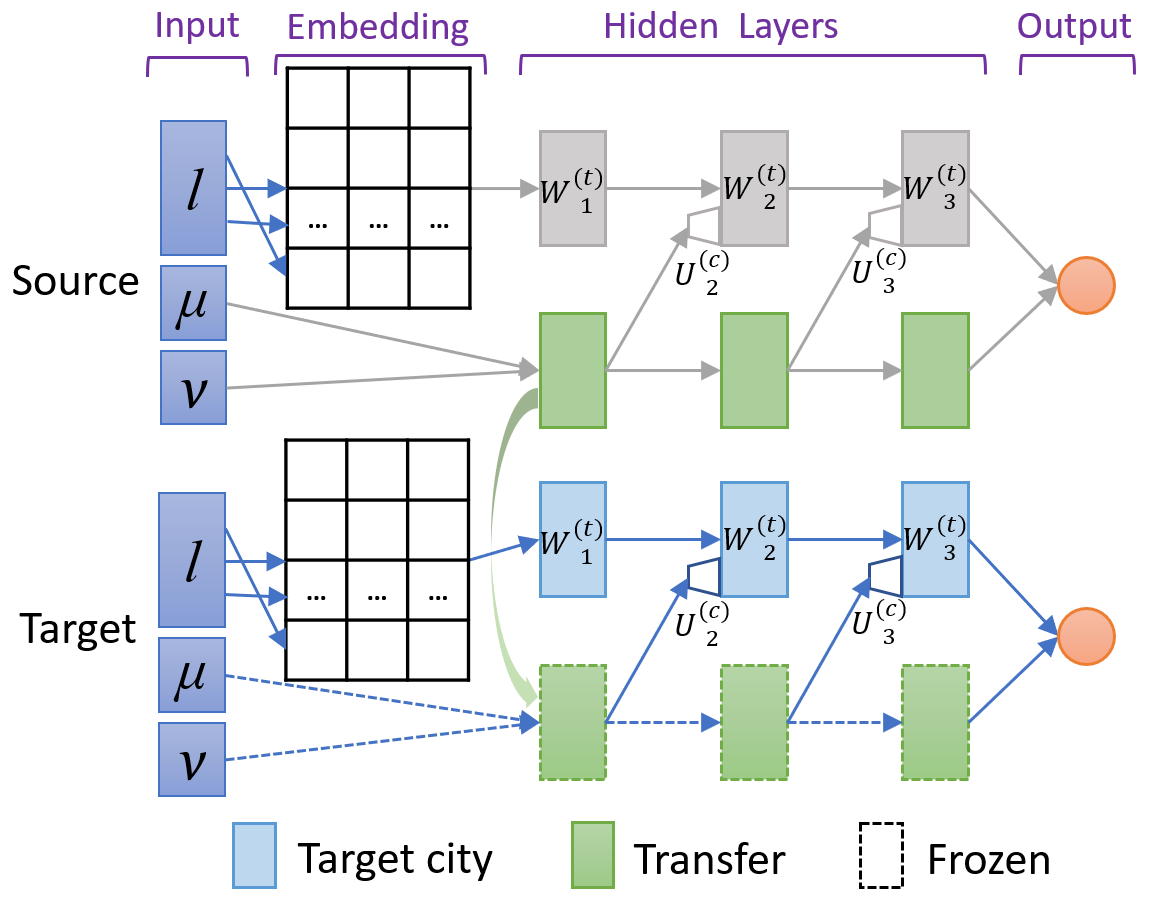}\hspace*{-0in}
    \caption{For CFPT network, we separate the input space and green blocks are the data tunnel that would process those features suitable for transfer. }
    \label{fig:cfpt_structure}
\end{center}
\end{figure}

\textbf{CFPT}: Different from the first two methods, CFPT (Figure~\ref{fig:cfpt_structure}) already separates data ``tunnels'' during the training on the source city. Green block only process correlated features between cities, which are suitable for transfer. After training on the source city, we copy the green blocks to the target network ( using the same structure). The lateral connection remains~\eqref{func:progressive_connection}. Notice this structure is unique because for the first two transfer methods, the source city uses the original CVNet structure in Figure~\ref{fig:vnet_structure}.

\section{Order Dispatching Simulator}

\begin{figure}
\begin{center}
    \hspace*{-0.0in}\adjincludegraphics[width=0.9\columnwidth, trim={0 0 0 0}, clip]{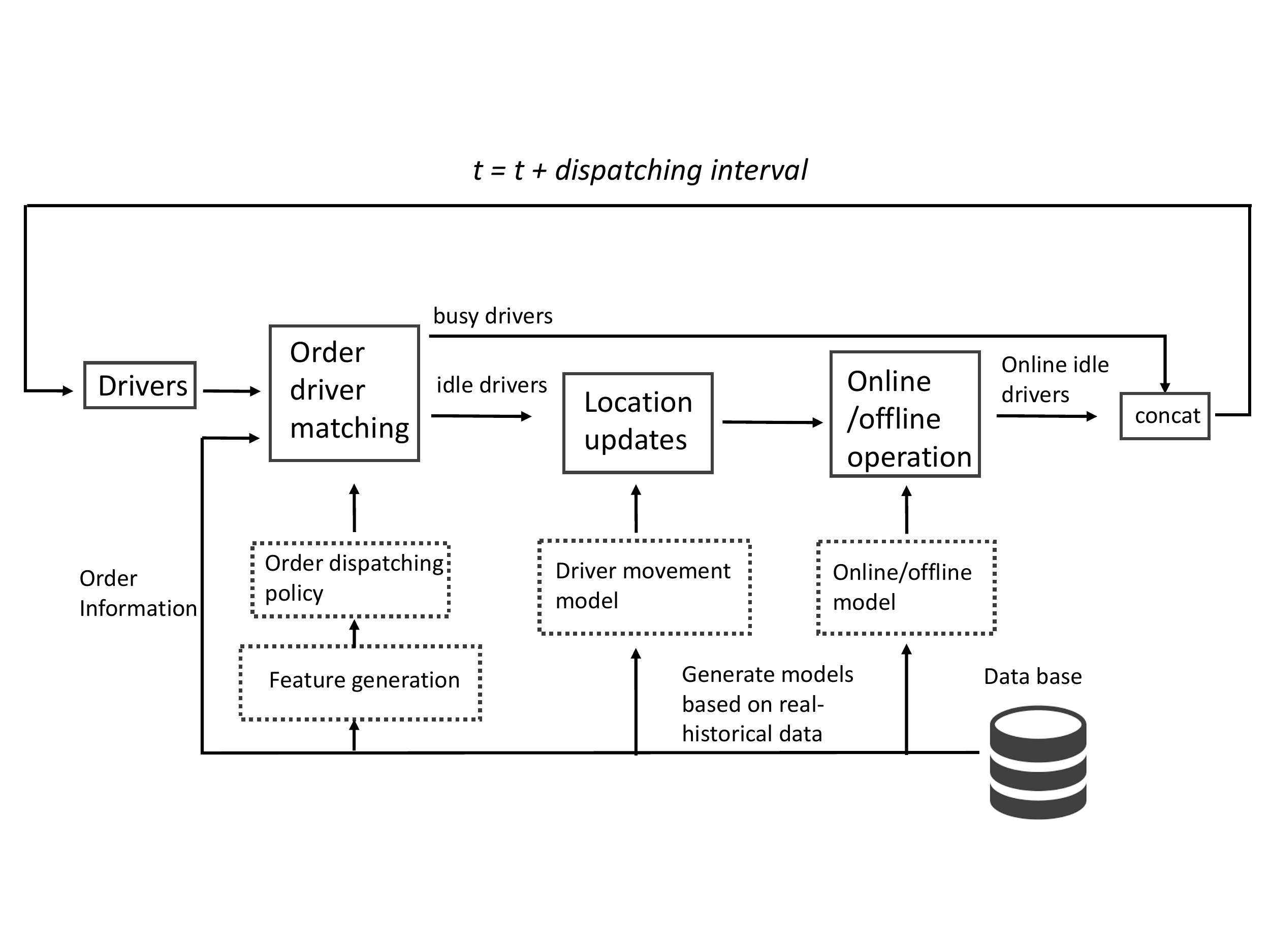}
    \caption{Composition and workflow of the order dispatching simulator.}
    \label{fig:simulator}
\end{center}
\end{figure}

We evaluate the proposed algorithms using a complicated and realistic order dispatching simulator. The simulator is able to provide an intuitive assessment for different order dispatching policies. Based on the historical real-data of particular dates, the simulator is first initialized by the the drivers' status as well as orders information of this date at the beginning of the simulation. Afterwards, the drivers' status are totally determined by the simulator, either fulfilling orders assigned by a certain order dispatching policy, or random walking followed by offline(online) operation according to certain models. Specifically, with the help of a particular order dispatching policy, the simulator periodically performs the order-driver matching using KM algorithm, where the drivers and orders form the bipartite graph. The meaning of the edge weights of the bipartite graph varies due to the difference of order dispatching policy (e.g., as to the distance-based policy, the weights between drivers and orders indicate the distance between orders' start location and drivers' location). Every time after order dispatching, drivers who assigned orders (busy drivers) would go the appointed locations, pick up passengers and carry passengres to the destination. Drivers who miss order assignment (idle drivers) would updates its destination according to a driver movement model. Moreover, before next order dispatching round, idle drivers could be offline and new drivers perhaps appear online in the platform. Therefore, an online/offline operation is performed according to the online/offline model. Both driver movement and the online/offline models are generated on the basis of real-world historical data. The workflow of the simulator is depicted in Fig~\ref{fig:simulator}.

\end{document}